\documentclass{article}

\PassOptionsToPackage{numbers, sort&compress}{natbib}

\usepackage[accepted]{icml2023}

\usepackage[utf8]{inputenc} %
\usepackage[T1]{fontenc}    %
\usepackage[hidelinks]{hyperref}  
\usepackage{url}            %
\usepackage{booktabs}       %
\usepackage{amsfonts}       %
\usepackage{nicefrac}       %
\usepackage{microtype}      %
\usepackage{xcolor}         %
\usepackage{multirow}
\usepackage[pdftex]{graphicx}
\usepackage{dsfont}
\usepackage{enumitem}
\usepackage{subfigure}
\usepackage{wrapfig}
\usepackage{float}
\usepackage{xr-hyper}

\makeatletter
\newcommand*{\addFileDependency}[1]{%
  \typeout{(#1)}
  \@addtofilelist{#1}
  \IfFileExists{#1}{}{\typeout{No file #1.}}
}
\makeatother

\usepackage{glossaries-extra}
\setabbreviationstyle[acronym]{long-short}
\glssetcategoryattribute{acronym}{nohyperfirst}{true}

\usepackage{xfrac}

\usepackage{amsmath,amsfonts,bm}

\def\eqref#1{equation~\ref{#1}}

\def\1{\bm{1}}

\def\vc{{\bm{c}}}

\def\vu{{\bm{u}}}

\def\vx{{\bm{x}}}
\def\vy{{\bm{y}}}

\DeclareMathAlphabet{\mathsfit}{\encodingdefault}{\sfdefault}{m}{sl}
\SetMathAlphabet{\mathsfit}{bold}{\encodingdefault}{\sfdefault}{bx}{n}

\def\gA{{\mathcal{A}}}

\def\gD{{\mathcal{D}}}

\def\gQ{{\mathcal{Q}}}

\def\gU{{\mathcal{U}}}

\newcommand{\E}{\mathbb{E}}

\newcommand{\softmax}{\mathrm{softmax}}

\DeclareMathOperator*{\argmax}{arg\,max}
\DeclareMathOperator*{\argmin}{arg\,min}

\usepackage[capitalize,noabbrev]{cleveref}
\usepackage{amsthm}
\newtheorem{theorem}{Theorem}

\def\Ln{L_0^\theta}
\def\La{L_1^\theta}
\def\Ia{y}
\def\I{\mathds{1}}
\newacronym{loe}{LOE}{latent outlier exposure}
\newacronym{aloe}{SOEL}{\textbf{s}emi-supervised \textbf{o}utlier \textbf{e}xposure with a \textbf{l}imited labeling budget}
\newacronym{ntl}{NTL}{neural transformation learning}
\newacronym{mhrot}{MHRot}{multi-head RotNet}
\newacronym{ad}{AD}{anomaly detection}

\def\s{S}

\def\kmeans{k-means++ }
\newcommand{\ditto}[1][.4pt]{superv. (\Cref{eqn:loss})}
\def\Margin{\text{Mar}}
\def\HybridA{\text{Hybr1}}
\def\PositiveB{\text{Pos1}}
\def\PositiveA{\text{Pos2}}
\def\RandomA{\text{Rand1}}
\def\RandomB{\text{Rand2}}
\def\HybridB{\text{Hybr2}}
\def\HybridC{\text{Hybr3}}
\Crefname{figure}{Fig.}{Figs.}
\Crefname{table}{Tab.}{Tabs.}
\Crefname{section}{Sec.}{Secs.}
\Crefname{proposition}{Prop.}{Props.}
\Crefname{equation}{Eq.}{Eqs.}
\Crefname{appendix}{Supp.}{Supps.}
\Crefname{theorem}{Thm.}{Thms.}
\Crefname{algorithm}{Alg.}{Algs.}

\usepackage{booktabs,array}

\newcount\rowc

\def\ttabular{%
\hbox\bgroup
\let\\\cr
\def\rulea{\ifnum\rowc=\@ne \hrule height 1.3pt \fi}
\def\ruleb{
\ifnum\rowc=1\hrule height 1.3pt \else
\ifnum\rowc=6\hrule height \heavyrulewidth 
   \else \hrule height \lightrulewidth\fi\fi}
\valign\bgroup
\global\rowc\@ne
\rulea
\hbox to 10em{\strut \hfill##\hfill}%
\ruleb
&&%
\global\advance\rowc\@ne
\hbox to 10em{\strut\hfill##\hfill}%
\ruleb
\cr}
\def\endttabular{%
\crcr\egroup\egroup}

\newcommand{\rbt}[1]{#1}

\icmltitlerunning{Deep Anomaly Detection under Labeling Budget Constraints}

\begin{document}

\twocolumn[
\icmltitle{Deep Anomaly Detection under Labeling Budget Constraints}

\icmlsetsymbol{equal}{*}
\icmlsetsymbol{advisor}{$\dag$}

\begin{icmlauthorlist}
\icmlauthor{Aodong Li}{equal,sch1}
\icmlauthor{Chen Qiu}{equal,comp}
\icmlauthor{Marius Kloft}{sch2}
\icmlauthor{Padhraic Smyth}{sch1}
\icmlauthor{Stephan Mandt}{advisor,sch1}
\icmlauthor{Maja Rudolph}{advisor,comp}
\end{icmlauthorlist}

\icmlaffiliation{sch1}{Department of Computer Science, University of California, Irvine, USA}
\icmlaffiliation{comp}{Bosch Center for Artificial Intelligence, Pittsburgh, USA}
\icmlaffiliation{sch2}{Department of Computer Science, TU Kaiserslautern, Germany}

\icmlcorrespondingauthor{Aodong Li}{aodongl1@uci.edu}
\icmlcorrespondingauthor{Stephan Mandt}{mandt@uci.edu}
\icmlcorrespondingauthor{Maja Rudolph}{Maja.Rudolph@us.bosch.com}

\icmlkeywords{Anomaly Detection, Active Learning, Semi-supervised Learning}

\vskip 0.3in
]

\printAffiliationsAndNotice{\icmlEqualContribution\icmlJointSupervision} %

\begin{abstract}
Selecting informative data points for expert feedback can significantly improve the performance of \gls{ad} in various contexts, such as medical diagnostics or fraud detection. In this paper, we determine a set of theoretical conditions under which anomaly scores generalize from labeled queries to unlabeled data. Motivated by these results, we propose a data labeling strategy with optimal data coverage under labeling budget constraints. In addition, we propose a new learning framework for semi-supervised \gls{ad}. 
Extensive experiments on image, tabular, and video data sets show that our approach results in state-of-the-art semi-supervised \gls{ad} performance under labeling budget constraints.
\end{abstract}

\section{Introduction}
Detecting anomalies in data is a fundamental task in machine learning with applications across multiple domains, from industrial fault detection to medical diagnosis. The main idea is to train a model (such as a neural network) on a data set of ``normal'' samples to minimize the loss of an auxiliary (e.g., self-supervised) task. Using the loss function to score test data, one hopes to obtain low scores for normal data and high scores for anomalies \citep{ruff2021unifying}. 

In practice, the training data is often contaminated with unlabeled anomalies that differ in unknown ways from the i.i.d. samples of normal data. 
No access to a binary anomaly label (indicating whether a sample is normal or not) makes learning the anomaly scoring function from contaminated data challenging; the training signal has to come exclusively from the input features (typically real-valued vectors).  Many approaches either assume that the unlabeled anomalies are too rarely encountered during training to affect learning \citep{wang2019effective} or try to detect and exploit the anomalies in the training data (e.g., \citet{qiu2022latent}).

\begin{figure*}[t!]
    \centering
    \includegraphics[width=0.95\linewidth]{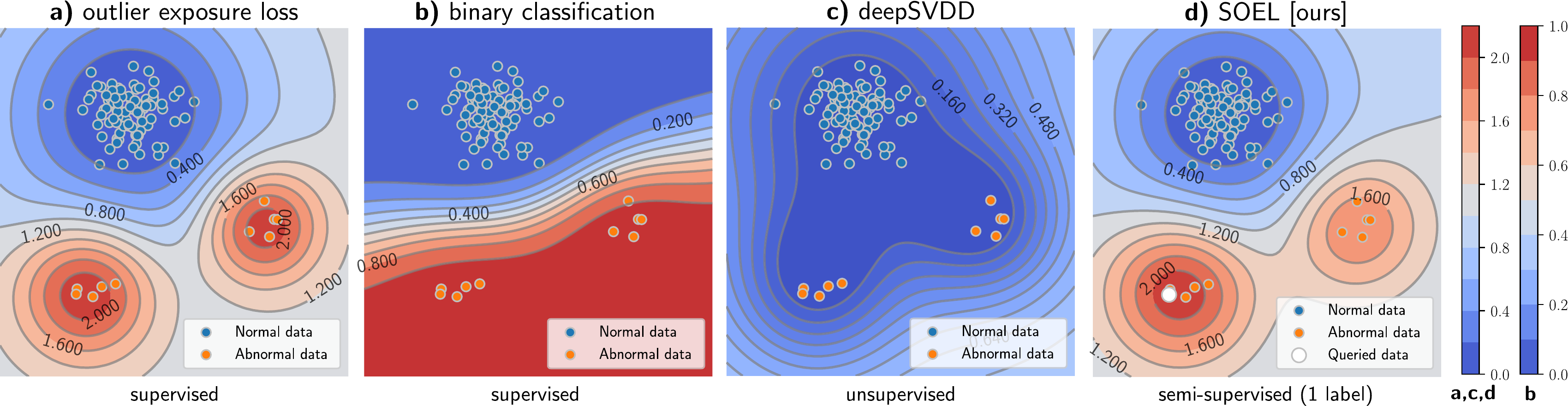}
    \vspace{-10pt}
    \caption{Anomaly score contour plots on 2D toy data demonstrate that \gls{aloe} [ours, (\textbf{d})] with only one labeled sample can achieve detection accuracy that is competitive with a fully supervised approach (\textbf{a}). Binary classification (\textbf{b}) is problematic for \gls{ad} since it cannot detect new anomalies, e.g. in the upper right corner of the plot. Subplot (\textbf{c}) demonstrates that unsupervised \gls{ad} is challenging with contaminated data. Even a single labeled query, in combination with our approach, can significantly improve \gls{ad}.}
    
    \label{fig:toy-data}
\vspace{-10pt}
\end{figure*}

While \gls{ad} is typically an unsupervised training task, sometimes expert feedback is available to check if individual samples are normal or not. For example, in a medical setting, one may ask a medical doctor to confirm whether a given image reflects normal or abnormal cellular tissue. Other application areas include detecting network intrusions or machine failures. Anomaly labels are usually expensive to obtain but are very valuable to guide an anomaly detector during training. For example, in \Cref{fig:toy-data}, we can see that our method, with only one labeled query (\Cref{fig:toy-data}~d) is almost on par with supervised \gls{ad} (\Cref{fig:toy-data}~a). However, the supervised setting is unrealistic, since expert feedback is typically expensive. Instead, it is essential to develop effective strategies for querying informative data points.

Previous work on \gls{ad} under a labeling budget primarily involves domain-specific applications and/or ad hoc architectures, making it hard to disentangle modeling choices from querying strategies \citep{trittenbach2021overview}. 
In contrast, this paper theoretically and empirically studies generalization performance using various labeling budgets, querying strategies, and losses. 

In summary, our main contributions are as follows:
\vspace{-7pt} %
\begin{enumerate}[leftmargin=*,itemsep=-2pt]
\item We prove that the ranking of anomaly scores generalizes from labeled queries to unlabeled data under certain conditions that characterize how well the queries cover the data. Based on this theory, we propose a diverse querying strategy for deep \gls{ad} under labeling budget constraints.

\item We propose \gls{aloe}, a semi-supervised learning framework  compatible with a large number of deep \gls{ad} losses. 
We show how all major hyperparameters can be %
eliminated, making \gls{aloe} easy to use.
To this end, we provide an estimate for the anomaly ratio in the data. 

\item We provide an extensive benchmark for deep \gls{ad} with a limited labeling budget. %
Our experiments on image, tabular, and video data provide evidence that \gls{aloe} outperforms existing methods significantly. Comprehensive ablations disentangle the benefits of each component.

\end{enumerate}

Our paper is structured as follows. \Cref{sec:methods} introduces  the problem setting we address and our main algorithm. \Cref{sec:related} discusses related work in deep \gls{ad}. \Cref{sec:experiments} discusses experimental results on each of image, video, and tabular data. Finally, we conclude this work in Section \ref{sec:discussion}.

\section{Methods}
\label{sec:methods}

\subsection{Notation and Problem Statement}%
\label{sec:prob-setup}
Consider a dataset $\{\vx_i\}_{i=1}^N$ where the datapoints $\vx_i$ are i.i.d. samples from a mixture distribution  $p(\vx)=(1-\alpha)p_0(\vx) + \alpha p_1(\vx)$. The distribution $p_0(\vx)$ corresponds to the normal data, while $p_1(\vx)$ corresponds to anomalous data. We assume that $0 \leq \alpha < 0.5$, i.e., that the anomalous data is non-dominant in the mixture; in practice, $\alpha \ll 0.5$. 

In the \gls{ad} problem, we wish to use the data to train an anomaly detector in the form of a parametric anomaly score function $\s(\vx;\theta)$. Once trained this score function is thresholded to determine whether a datapoint $\vx_i$ is anomalous, as indicated by the binary anomaly label  $y_i := y(\vx_i) \in \{0 := \text{``normal''}, 1:=\text{``abnormal''}\}$.

We focus on the situation where the training data is unlabeled (only $\vx_i$ is known, not $y_i$), but where we have access to an oracle (e.g., a human expert) that is able to provide labels $y_i$ for a budgeted number $K$ of the $N$ training points.

\subsection{Outline of the Technical Approach}

Our work addresses the following questions: How to best select informative data points for labeling -- this is called the {\em querying strategy}, how to best learn an anomaly detector from both the labeled and unlabeled 
data in a semi-supervised fashion, and how to make the approach easy to use by eliminating a crucial hyper-parameter.

\textbf{Querying Strategy.} A successful approach for deep \gls{ad} under labeling budget constraints will require a strategy for selecting the most beneficial set of queries. We choose a theoretically-grounded approach based on generalization performance. For this, we exploit that at test-time an \gls{ad} method will threshold the anomaly scores to distinguish between normal samples and anomalies. This means that the quality of a scoring function is not determined by the absolute anomaly scores but only by their relative ranking. 
In \Cref{sec:active-learning}, we characterize a favorable property of the query set which can guarantee that the ranking of anomaly scores generalizes from the labeled data to unlabeled samples. Since this is desirable, we derive a querying strategy that under a limited labeling budget best fulfills the favorable properties put forth by our analysis.

\textbf{Semi-supervised Outlier Exposure.} As a second contribution, we propose a semi-supervised learning framework that best exploits both the labeled query set and the unlabeled data. It builds on supervised \gls{ad} and \gls{loe} which we review in \Cref{sec:background}. 
We present \gls{aloe} in \Cref{sec:aloe-loss}. The \gls{aloe} training objective is designed to receive opposing training signals from the normal samples and the anomalies. An EM-style algorithm alternates between estimating the anomaly labels of the unlabaled data and improving the anomaly scoring function using the data samples and their given or estimated labels.

\textbf{Hyperparameter Elimination.} Like related methods discussed in \Cref{sec:related}, \gls{aloe} has an important hyperparameter $\alpha$ which corresponds to the expected fraction of anomalies in the data. While previous work has to assume that $\alpha$ is known \citep{qiu2022latent}, our proposed method presents an opportunity to estimate it. 
The estimate has to account for the fact that the optimal querying strategy derived from our theory in \Cref{sec:active-learning} is not i.i.d.. In \Cref{sec:alpha}, we provide an estimate of $\alpha$ for any stochastic querying strategy.

\subsection{Background: Deep \gls{ad}}
\label{sec:background}
In deep \gls{ad}, 
 auxiliary losses help learn the anomaly scoring function $\s(\vx;\theta)$. Popular losses include autoencoder-based losses \citep{zhou2017anomaly}, the deep SVDD loss \citep{ruff2018deep}, or the neural transformation learning loss \citep{qiu2021neural}. It is assumed that minimizing such a 
 loss $L^\theta_0(\vx)\equiv {\cal L}_0(\s(\vx;\theta))$  over ``normal'' data leads to a desirable scoring function that assigns low scores to normal samples and high scores to anomalies. 

Most deep \gls{ad} methods optimize such an objective over an entire unlabeled data set, even if it contains unknown anomalies. It is assumed that the anomalies are rare enough that they will not dilute the training signal provided by the normal samples ({\em inlier priority}, \citep{wang2019effective}). Building on the ideas of \citet{ruff2019deep} that synthetic anomalies can provide valuable training signal, \citet{qiu2022latent} show how to discover and exploit anomalies by treating the anomaly labels as latent variables in training.

The key idea of \citet{ruff2019deep} is to construct a complementary loss $L^\theta_1(\vx)\equiv {\cal L}_1(\s(\vx;\theta))$ for anomalies that has an opposing effect to the normal loss $L^\theta_0(\vx)$. For example, the deep SVDD loss $L_0^\theta(\vx)=||f_\theta(\vx)-{\bf c}||^2$, with feature extractor $f_\theta$, pulls normal data points towards a fixed center $c$ \citep{ruff2018deep}. The opposing loss for anomalies, defined in \citet{ruff2019deep} as $L_1^\theta(\vx) = 1/L_0^\theta(\vx)$, pushes abnormal data away from the center.

\textbf{Supervised \gls{ad}.} Using only the labeled data indexed by $\gQ$ one could train $\s(\vx;\theta)$ using a supervised loss \citep{hendrycks2018deep,gornitz2013toward}
\begin{equation}
\label{eqn:loss}
 {\cal L}_\gQ(\theta) = \frac{1}{|\gQ|}\sum_{j\in \gQ} \big(y_j \La(\vx_j) + (1-y_j) \Ln(\vx_j) \big) .
\end{equation}
\textbf{Latent Outlier Exposure.}
Latent outlier exposure (LOE, \citep{qiu2022latent}) is an unsupervised \gls{ad} framework that uses the same loss as \Cref{eqn:loss} but treats the labels $y$ as latent variables. An EM-style algorithm alternates between optimizing the model w.r.t. $\theta$ and inferring the labels $y$.

In this work, we propose semi-supervised outlier exposure with a limited labeling budget (SOEL) which builds on these ideas. We next present the querying strategy and \rbt{when the querying strategy leads to correct ranking of anomaly scores (\Cref{sec:active-learning}),} the \gls{aloe} loss (\Cref{sec:aloe-loss}), and how the hyperparameter $\alpha$ can be eliminated (\Cref{sec:alpha})

\subsection{Querying Strategies for \gls{ad}} 
\label{sec:active-learning}

The first ingredient of \gls{aloe} is a querying strategy for selecting informative data points to be labeled,
which we derive from theoretical considerations. %
An important property of the querying strategy is how well it covers unlabeled data. The quality of a querying strategy is determined by the smallest radius $\delta$ such that all unlabeled points are within distance $\delta$ of one queried sample of the same type. In this paper, we prove that if the queries cover both the normal data and the anomalies well (i.e., if $\delta$ is small), a learned anomaly detector that satisfies certain conditions is guaranteed to generalize correctly to the unlabeled data (The exact statement and its conditions will be provided in \Cref{thm:rank}). Based on this insight, we propose to use a querying strategy that is better suited for deep \gls{ad} than previous work.

\begin{theorem}
\label{thm:rank}
Let $\gQ_0$ be the index set of datapoints labeled normal and $\gQ_1$ the index set of datapoints labeled abnormal. 
Let $\delta \in \mathbb{R}^+$ be the smallest radius, such that for each unlabeled anomaly $\vu_a$ and each unlabeled normal datum $\vu_n$ there exist labeled data points $\vx_a, a\in\gQ_1$ and $\vx_n, n\in\gQ_0$, such that $\vu_a$ is within the $\delta$-ball of $\vx_a$ and $\vu_n$ is within the $\delta$-ball around  $\vx_n$. If a $\lambda_s$-Lipschitz continuous function $\s$ ranks the labeled data correctly, with a large enough margin, i.e. $\s(\vx_a)-\s(\vx_n) \geq 2\delta\lambda_s$,
then $\s$ ranks the unlabeled points correctly, too, and $\s(\vu_a)%
\geq\s(\vu_n)$.
\end{theorem}

In \Cref{app:thm1}, we prove \Cref{thm:rank} and discuss the assumptions. An implication of this theorem is that a smaller $\delta$ corresponding to a tighter cover of the data leads to better-generalized ranking performance. As detailed in \Cref{app:thm1}, there is a connection between correct anomaly score ranking and high AUROC performance, a common evaluation metric for \gls{ad}.

Existing methods use querying strategies that do not have good coverage and are therefore not optimal under \Cref{thm:rank}. For a limited querying budget, random querying puts too much weight on high-density areas of the data space, while other strategies only query locally, e.g., close to an estimated decision boundary between normal and abnormal data. 

\textbf{Proposed Querying Strategy.} Based on \Cref{thm:rank}, we propose a querying strategy that encourages tight coverage: diverse querying.
In practice, we use the seeding algorithm of \kmeans which is usually used to initialize diverse clusters.\footnote{This has complexity $O(KN)$ which can be reduced to $O(K\log N)$ using scalable alternatives \citep{bahmani2012scalable}.}  
It iteratively samples another data point to be added to the query set $\gQ$ until the labeling budget is reached. Given the existing queried samples, the probability of drawing another query from the unlabeled set $\gU$ is proportional to its distance to the closest sample already in the query set $\gQ$:
\begin{align}
\label{eq:query-prob}
    p_{\rm query} (\vx_i) = \softmax\big(h(\vx_i)/\tau\big) \quad \forall i \in \gU,
\end{align}
The temperature parameter $\tau$ controls the diversity of the sampling procedure, and  
$h(\vx_i) = \min_{\vx_j \in \mathcal{Q}} d(\vx_i,\vx_j)$ is the distance of a sample $\vx_i$ to the query set $\mathcal{Q}$.
For a meaningful notion of distance, we define $d$ in an embedding space as $d(\vx,\vx') = \|\phi(\vx)-\phi(\vx')\|_2$, where $\phi$ is a neural feature map. We stress that all deep methods considered in this paper have an associated feature map that we can use.
The fact that L2 distance is used in the querying strategy is not an ad-hoc choice but rather aligned with the $\delta$-ball radius definition (\Cref{eq:delta-est} in \Cref{app:thm1}) in \Cref{thm:rank}. 

In \Cref{app:thm1}, we discuss the cover radius and empirically validate that diverse querying leads to smaller $\delta$ than others and is hence advantageous for \gls{ad}.

\subsection{Semi-supervised Outlier Exposure Loss (SOEL)}
\label{sec:aloe-loss}

We next consider how to use both labeled and unlabeled samples in training. We propose \gls{aloe} whose loss combines the unsupervised \gls{ad} loss of LOE~\citep{qiu2022latent} for the unlabeled data with the supervised loss (\Cref{eqn:loss}) for the labeled samples. 
For all queried data (with index set $\gQ$), we assume that ground truth labels $y_i$  are available, while for unqueried data (with index set $\gU$), the labels $\tilde{y}_i$ are unknown. Adding both losses together yields
\begin{align}
\label{eqn:loss-2}
 {\cal L}(\theta, \tilde{\vy}) = \frac{1}{|\gQ|}\sum_{j\in \gQ} \big(y_j \La(\vx_j) + (1-y_j) \Ln(\vx_j) \big) + \nonumber \\ 
 \frac{1}{|\gU|}\sum_{i\in \gU} \big(\tilde{y}_i \La(\vx_i) + (1-\tilde{y}_i) \Ln(\vx_i)\big).
\end{align}
Similar to \citet{qiu2022latent}, optimizing this loss involves a block coordinate ascent scheme that alternates between inferring the unknown labels and taking gradient steps to minimize \Cref{eqn:loss-2} with the inferred labels. In each iteration, the pseudo labels $\tilde{y}_i$ for $i\in \gU$ are obtained by minimizing \Cref{eqn:loss-2} subject to a constraint of $\sum_{i\in\gQ} y_i + \sum_{i\in\gU} \tilde{y}_i = \alpha N$.     
The constraint ensures that the inferred anomaly labels respect a certain contamination ratio $\alpha$. To be specific, let $\tilde\alpha$ denote the fraction of anomalies among the \emph{unqueried} set $\gU$, so that 
$\tilde\alpha |\gU| + \sum_{j\in\gQ}y_j =\alpha N$. 
The constrained optimization problem is then solved by using the current anomaly score function $S$ to rank the unlabeled samples and assign the top $\tilde\alpha$-quantile of the associated labels $\tilde y_i$ to the value $1$, and the remaining to the value $0$. 

We illustrate \gls{aloe}'s effect on a 2D toy data example in \Cref{fig:toy-data}, where \gls{aloe} (\textbf{d}) almost achieves the same performance as the supervised AD (\textbf{c}) with only one queried point.   %

In theory, $\alpha$ could be treated as a hyperparameter, but eliminating hyperparameters is important in \gls{ad}. In many practical applications of \gls{ad}, there is no labeled data that can be used for validation. While \citet{qiu2022latent} have to assume that the contamination ratio is given, \gls{aloe} provides an opportunity to estimate $\alpha$. In \Cref{sec:alpha}, we develop an importance-sampling based approach to estimate $\alpha$ from the labeled data. \rbt{Estimating this ratio can be beneficial for many \gls{ad} algorithms, including OC-SVM~\citep{scholkopf2001estimating}, kNN~\citep{ramaswamy2000efficient}, Robust PCA/Auto-encoder~\citep{zhou2017anomaly}, and Soft-boundary deep SVDD~\citep{ruff2018deep}. When working with contaminated data, these algorithms require a decent estimate of the contamination ratio for good performance.} %

Another noteworthy aspect of the \gls{aloe} loss is that it weighs the \emph{averaged} losses equally to each other. In \Cref{app:ablation_weight}, we empirically show that equal weighting yields the best results among a large range of various weights. This provides more weight to every queried data point than to an unqueried one, because we expect the labeled samples to be more informative. On the other hand, it ensures that neither loss component will dominate the learning task. Our equal weighting scheme is also practical because it avoids a hyperparameter.

\subsection{Contamination Ratio Estimation.}
\label{sec:alpha}
To eliminate a critical hyperparameter in our approach, we estimate the \emph{contamination ratio} $\alpha$, i.e., the fraction of anomalies in the dataset.  Under a few assumptions, we show how to estimate this parameter using mini-batches composed of on non-i.i.d. samples. 

We consider the contamination ratio $\alpha$ as the fraction of anomalies in the data. We draw on the notation from~\Cref{sec:prob-setup} to define $\Ia(\vx)$ as an oracle, outputting $1$ if $\vx$ is an anomaly, and $0$ otherwise (e.g., upon \emph{querying} $\vx$). We can now write $\alpha = \mathbb{E}_{p(\vx)} [\Ia(\vx)]$. 

Estimating $\alpha$ would be trivial given an unlimited querying budget of i.i.d. data samples. The difficulty arises due to the fact that (1) our querying budget is limited, and (2) we query data in a non-i.i.d. fashion so that the sample average is not representative of the anomaly ratio of the full data set. 

Since the queried data points are not independently sampled, we cannot straightforwardly estimate $\alpha$ based on the empirical frequency of anomalies in the query $\gQ$. More precisely, our querying procedure results in a  chain of indices $\gQ = \{i_1, i_2, ..., i_{|\gQ|}\}$, where $i_1 \sim \rm{Unif}(1:N)$, and each conditional distribution $ i_k|i_{<k}$ is defined by \Cref{eq:query-prob}. We will show as follows that this sampling bias can be compensated using importance weights. 
 
As follows, we first propose an importance-weighted estimator of $\alpha$ and then prove the estimator is unbiased under certain idealized conditions specified by two assumptions about our querying strategy. Justifications for the two assumptions will be provided below.

For a random query $\gQ$, its anomaly scores $\{S(\vx_i): i\in\gQ\}$ and anomaly labels $\{\Ia(\vx_i):i\in\gQ\}$ are known. Write $S(\vx_i)$ as $s_i$ and let $p_s(s_i)$ denote the marginal density of population anomaly scores and $q_s(s_i)$ denote the marginal density of the queried samples' anomaly scores. Our importance-weighted estimator of the contamination ratio is
\begin{align}
\label{eq:alpha-is}
    \hat{\alpha} 
    = \frac{1}{|\gQ|}\sum_{i=1}^{|\gQ|}\frac{p_s(s_i)}{q_s(s_i)}\Ia(\vx_i). 
\end{align}
As discussed above, $\Ia(\vx_i)$ are the ground truth anomaly labels, obtained from querying $\gQ$. The estimator takes into account that, upon repulsive sampling, we will sample data points in the tail regions of the data distribution more often than we would upon uniform sampling. 

In practice, we learn $p_s$ and $q_s$ using a kernel density estimator in the one-dimensional space of anomaly scores of the training data and the queried data, respectively. We set the bandwidth to the average spacing of scores. With the following two assumptions, \Cref{eq:alpha-is} is unbiased.

\textbf{Assumption 1.} \textsl{The anomaly scores $\{S(\vx_i): i \in \gQ\}$ in a query set $\gQ$ are approximately independently distributed.
}

\textbf{Assumption 2.} \textsl{
Let $\Ia_s(S(\vx))$ denote an oracle that assigns ground truth anomaly labels based on the model's anomaly scores $S(\vx)$.
We assume that such an oracle exists, i.e., the anomaly score $S(\vx)$ is a sufficient statistic of the ground truth anomaly labeling function: $\Ia_s(S(\vx)) = \Ia(\vx)$.}

Assumptions 1 and 2 are only approximations of reality. In our experiment section, we will show that they are good working assumptions to estimate anomaly ratios. Below, we will provide additional strong evidence that assumptions 1 and 2 are well justified. 

The following theorem is a consequence of them:

\begin{theorem}
\label{thm:alpha}
Assume that Assumptions 1 and 2 hold. Then, \Cref{eq:alpha-is} is an unbiased estimator of the contamination ratio $\alpha$, i.e., $\E[\hat\alpha]=\alpha$.

\end{theorem}

\rbt{The proof is in \Cref{sec:assum-veri}.} 
\Cref{thm:alpha} allows us to estimate the contamination ratio based on a non-iid query set $\gQ$.

\textbf{Discussion.} We empirically verified the fact that \Cref{thm:alpha} results in reliable estimates for varying contamination ratios 
in \Cref{app:alpha-est}. Since Assumptions 1 and 2 seem strong, we discuss their justifications and empirical validity next. 

While verifying the independence assumption (Assumption 1) rigorously is difficult, we tested for linear correlations between the scores (\Cref{sec:assum-veri-1}). We found that the absolute off-diagonal coefficient values are significantly smaller than one on CIFAR-10, providing support for Assumption 1. 
A heuristic argument can be provided to support the validity of Assumption 1 based on the following intuition. When data points are sampled diversely in a high-dimensional space, the negative correlations induced by their repulsive nature tend to diminish when the data is projected onto a one-dimensional subspace. This intuition stems from the fact that a high-dimensional ambient space offers ample dimensions for the data points to avoid clustering. To illustrate this, consider the scenario of sampling diverse locations on the Earth's surface, with each location representing a point in the high-dimensional space. By including points from various continents, we ensure diversity in their spatial distribution. However, when focusing solely on the altitude of these locations (such as distinguishing between mountain tops and flat land), it is plausible that the altitude levels are completely uncorrelated. While this heuristic argument provides an intuitive understanding, it is important to note that it does not offer a rigorous mathematical proof.

To test Assumption 2, we tested the degree to which the anomaly score is a sufficient statistic for anomaly scoring on the training set. The assumption would be violated if we could find pairs of training data $\vx_i$ and $\vx_j$, where $\vx_i\neq\vx_j$, with identical anomaly scores $S(\vx_i)=S(\vx_j)$ but different anomaly labels $\Ia_s(s_i)\neq \Ia_s(s_j)$\footnote{The condition $S(\vx_i)\neq S(\vx_j)$ for $\vx_i\neq\vx_j$ hints we can assign a unique label to each data point based on their scores.}. On FMNIST, we found 38 data pairs with matching scores, and none of them had opposite anomaly labels. For CIFAR-10, the numbers were 21 and 3, respectively. See \Cref{sec:assum-veri-3} for details.

\section{Related Work}
\label{sec:related}

\textbf{Deep Anomaly Detection.}
Many recent advances in anomaly detection are in the area of deep learning \citep{ruff2021unifying}. One early strategy was to use autoencoder- \citep{principi2017acoustic,zhou2017anomaly} or density-based models \citep{schlegl2017unsupervised, deecke2018image}. Another pioneering stream of research combines one-class classification \citep{scholkopf2001estimating} with deep learning for unsupervised \citep{ruff2018deep,ijcai2022p305} and semi-supervised \citep{ruff2019deep} anomaly detection. Many other approaches to deep anomaly detection are self-supervised. %
They employ a self-supervised loss function to train the detector and score anomalies \citep{golan2018deep,hendrycks2019using,bergman2020classification,qiu2021neural,shenkar2022anomaly,schneider2022detecting}. Our work resides in the self-supervised anomaly detection category and can be extended to other data modalities if an appropriate loss is provided.%

While all these methods assume that the training data consists of only normal samples, in many practical applications, the training pool may be contaminated with unidentified anomalies \citep{vilhjalmsson2013nature,poisoning1}.
This can be problematic because the detection accuracy typically deteriorates when the contamination ratio increases \citep{wang2019effective}. Addressing this, refinement \citep{zhou2017anomaly,yoon2021self} attempts to cleanse the training pool by removing anomalies therein, although they may provide valuable training signals. As a remedy, \citet{qiu2022latent} propose to jointly infer binary labels to each datum (normal vs. anomalous) while updating the model parameters based on outlier exposure. Our work also makes the contaminated data assumption and employs the training signal of abnormal data.

\textbf{Querying Strategies for Anomaly Detection.} Querying strategies play an important role in batch active learning \citep{sener2018active, ash2020deep, citovsky2021batch, pinsler2019bayesian, hoi2006large} but are less studied for anomaly detection.
The human-in-the-loop setup for anomaly detection has been pioneered by \citet{pelleg2004active}. %
Query samples are typically chosen locally, e.g., close to the decision boundary of a one-class SVM \citep{gornitz2013toward,yin2018active} or sampled according to a density model \citep{ghasemi2011active}. \citet{siddiqui2018feedback,das2016incorporating} propose to query the most anomalous instance, while \citet{das2019active} employ a tree-based ensemble to query both anomalous and diverse samples. A recent survey compares various aforementioned query strategies with one-class classifiers \citep{trittenbach2021overview}.

\citet{pimentel2020deep} query samples with the top anomaly scores for autoencoder-based methods, while \citet{ning2022deep} improve the querying by considering the diversity. \citet{tang2020deep} use an ensemble of deep anomaly detectors and query the most likely anomalies for each detector separately. \citet{russo2020active} query samples where the model is uncertain about the predictions. \citet{pang2021toward} and \citet{zha2020meta} propose querying strategies based on reinforcement learning, which requires labeled datasets. 

All these querying strategies do not optimize coverage as defined in \Cref{thm:rank}, and as a result, their generalization guarantees are less favorable than our method.   
Most querying strategies from the papers discussed above are fairly general and can be applied in combination with various backbone models. 
Since more powerful backbone models have been released since these earlier publications, we ensure a fair comparison by studying all querying strategies in combination with the same backbone models as \gls{aloe}.

\section{Experiments}
\label{sec:experiments}

\begin{table*}[t!]
\vspace{-5pt}
	\caption{A summary of all compared experimental methods' query strategy and training strategy irrespective of their backbone models.}
	\label{tab:baselines}
 	\footnotesize
	\centering
	\begin{tabular}{lll@{\hskip 0.2in}cc}
        \toprule
		Name & Reference & Querying Strategy & Loss (labeled) & Loss (unlabeled)  \\
		\midrule
		\Margin & \citet{gornitz2013toward}  &margin query %
		& superv. (\Cref{eqn:loss}) & one class \\
        \HybridA & \citet{gornitz2013toward} & margin diverse query %
        & \ditto & one class  \\
        \PositiveB & \citet{pimentel2020deep} & most positive query & \ditto & none \\
        \PositiveA & \citet{barnabe2015active} & most positive query & \ditto & one class \\
        
        \RandomA & \citet{ruff2019deep} & random query & \ditto & one class  \\
        \RandomB & \citet{trittenbach2021overview} & positive random query & \ditto & one class    \\
        \HybridB &\citet{das2019active} &positive diverse query & \ditto & none \\
        \HybridC & \citet{ning2022deep} &positive diverse query & refinement & weighted one class \\
        \midrule
        SOEL& [ours] & diverse (\Cref{eq:query-prob}) & \multicolumn{2}{c}{ semi-supervised outlier exposure loss (\Cref{eqn:loss-2})} \\
        \bottomrule
	\end{tabular}
 	\vspace{-5pt}
\end{table*}
We study \gls{aloe} on standard image benchmarks, medical images, tabular data, and surveillance videos.
Our extensive empirical study establishes how our proposed method compares to eight \gls{ad} methods with labeling budgets implemented as baselines.
We first describe the baselines and their implementations (\Cref{tab:baselines}) and then the experiments on images (\Cref{sec:image_exp}), tabular data (\Cref{sec:tab_exp}), videos (\Cref{sec:video_exp}) and finally additional experiments (\Cref{sec:add-exp}). %

\begin{table*}[t!]
\vspace{-5pt}
    \caption{AUC ($\%$) with standard deviation for anomaly detection on 11 image datasets when the query budget $|\gQ|=20$. \gls{aloe} outperforms all baselines by a large margin by querying diverse and informative samples.}%
\label{tab:img_results}
    \footnotesize
    \centering
  \begin{tabular}{l|ccccccccc}
  \toprule
     & \bfseries\Margin & \bfseries\HybridA & \bfseries\PositiveB & \bfseries\PositiveA & \bfseries\RandomA & \bfseries\RandomB & \bfseries\HybridB & \bfseries\HybridC & \bfseries SOEL \\ 
  \midrule
    \bfseries CIFAR10                  & 92.4$\pm$0.7                                    & 92.0$\pm$0.7                                     & 93.4$\pm$0.5                                       & 92.1$\pm$0.7                                       & 89.2$\pm$3.2                                     & 91.4$\pm$1.0                                     & 85.1$\pm$2.2                                     & 71.8$\pm$7.4                                     & \textbf{96.3$\pm$0.3}                  \\ 
\bfseries FMNIST                   & 93.1$\pm$0.4                                    & 92.6$\pm$0.4                                     & 92.2$\pm$0.6                                       & 89.3$\pm$1.0                                       & 84.0$\pm$3.8                                     & 90.6$\pm$1.1                                     & 88.7$\pm$1.4                                     & 82.6$\pm$4.3                                     & \textbf{94.8$\pm$0.6}                  \\ 
\bfseries Blood                    & 68.6$\pm$1.8                                    & 69.1$\pm$1.3                                     & 69.6$\pm$1.8                                       & 72.2$\pm$4.9                                       & 70.6$\pm$1.6                                     & 69.2$\pm$1.7                                     & 72.2$\pm$2.7                                     & 58.3$\pm$5.2                                     & \textbf{80.5$\pm$0.5}                  \\ 
\bfseries OrganA                   & 86.4$\pm$1.3                                    & 87.4$\pm$0.7                                     & 81.7$\pm$2.9                                       & 81.8$\pm$2.1                                       & 82.9$\pm$0.6                                     & 86.5$\pm$0.7                                     & 88.6$\pm$1.5                                     & 68.8$\pm$3.1                                     & \textbf{90.7$\pm$0.7}                  \\ 
\bfseries OrganC                   & 86.5$\pm$0.9                                    & 87.0$\pm$0.7                                     & 84.6$\pm$1.9                                       & 79.6$\pm$2.0                                       & 85.5$\pm$0.9                                     & 86.4$\pm$0.8                                     & 84.8$\pm$1.2                                     & 68.9$\pm$3.0                                     & \textbf{89.7$\pm$0.7}                  \\ 
\bfseries OrganS                   & 83.5$\pm$1.1                                    & 84.1$\pm$0.4                                     & 83.2$\pm$1.3                                       & 78.6$\pm$1.0                                       & 82.2$\pm$1.4                                     & 83.8$\pm$0.4                                     & 82.3$\pm$0.7                                     & 66.9$\pm$4.3                                     & \textbf{87.4$\pm$0.8}                  \\ 
\bfseries OCT                      & 64.4$\pm$3.7                                    & 63.3$\pm$1.8                                     & 63.8$\pm$4.4                                       & 63.0$\pm$4.0                                       & 59.7$\pm$1.9                                     & 62.1$\pm$4.3                                     & 63.0$\pm$7.6                                     & 56.2$\pm$4.5                                     & \textbf{68.5$\pm$3.4}                  \\ 
\bfseries Path                     & 82.7$\pm$2.4                                    & 86.0$\pm$1.1                                     & 77.5$\pm$2.0                                       & 80.2$\pm$3.5                                       & 83.2$\pm$1.6                                     & 83.9$\pm$2.9                                     & 86.1$\pm$2.0                                     & 75.1$\pm$4.2                                     & \textbf{88.1$\pm$1.1}                  \\ 
\bfseries Pneumonia                & 72.1$\pm$7.0                                    & 75.1$\pm$5.3                                     & 75.5$\pm$8.8                                       & 83.6$\pm$6.1                                       & 68.1$\pm$5.9                                     & 76.0$\pm$8.0                                     & 88.4$\pm$3.3                                     & 63.4$\pm$17.7                                    & \textbf{91.2$\pm$1.4}                  \\ 
\bfseries Tissue                   & 60.2$\pm$1.5                                    & 61.3$\pm$1.7                                     & 65.8$\pm$1.7                                       & 63.5$\pm$2.0                                       & 59.9$\pm$1.7                                     & 59.5$\pm$1.3                                     & 62.1$\pm$1.7                                     & 50.8$\pm$1.6                                     & \textbf{66.4$\pm$1.4}                  \\ 
\bfseries Derma                    & 62.6$\pm$3.8                                    & 63.1$\pm$4.7                                     & 66.6$\pm$2.3                                       & 66.4$\pm$4.3                                       & 64.5$\pm$4.8                                     & 68.3$\pm$2.1                                     & 57.2$\pm$13.3                                    & 48.0$\pm$13.6                                    & \textbf{73.5$\pm$2.5}                  \\ 
\midrule
\bfseries Average & 77.5 & 78.3 &  77.3  & 77.6 &75.4 &78.0 &78.0 &64.6 &\textbf{84.3} \\
  \bottomrule 
  \end{tabular}
 
\vspace{-5pt}
\end{table*}
\textbf{Baselines.} 
Most existing baselines apply their proposed querying and training strategies to shallow \gls{ad} methods or sub-optimal deep models (e.g., autoencoders \citep{zhou2017anomaly}).
In recent years, these approaches have consistently been outperformed by self-supervised \gls{ad} methods \citep{hendrycks2019using}. For a fair comparison, we endow all baselines with the same self-supervised backbone models also used in our method. %
By default we use \gls{ntl} \citep{qiu2021neural} as the backbone model, which was identified as state-of-the-art in a recent independent comparison of 13 models \citep{alvarez2022revealing}. Results with other backbone models are shown in \Cref{app:ablation_model}. %

The baselines are summarized in \Cref{tab:baselines} and detailed in \Cref{sec:baselines}.  
They differ in their querying strategies (col. 3) and training strategies (col. 4 \& 5): the unlabeled data is either ignored or modeled with a one-class objective. 
Most baselines incorporate the labeled data by a supervised loss (\Cref{eqn:loss}).
As an exception, \citet{ning2022deep} remove all queried anomalies and then train a weighted one-class objective on the remaining data. All baselines weigh the unsupervised and supervised losses equally. 
They differ in their querying strategies, summarized below:
\vspace{-7pt}
\begin{itemize}[leftmargin=*,itemsep=-2pt]
    \item {\bf Margin query} selects samples close to the boundary of the normality region deterministically. The method uses the true contamination ratio to choose an ideal boundary.
    \item {\bf Margin diverse query} combines margin query with neighborhood-based diversification. 
    It selects samples that are not $k$-nearest neighbors of the queried set.
    Thus samples are both diverse and close to the boundary.
    \item {\bf Most positive query} always selects the top-ranked samples ordered by their anomaly scores.
    \item {\bf Positive diverse query} combines querying according to anomaly scores with distance-based diversification. 
    The selection criterion combines anomaly score and the minimum Euclidean distance to all queried samples.
    \item {\bf Random query} draws samples uniformly.  
    \item {\bf Positive random query} samples uniformly among the top $50\%$ data ranked by anomaly scores.  
\end{itemize}
\vspace{-2pt}

\textbf{Implementation Details.}
In all experiments, we use a \gls{ntl} \citep{qiu2021neural} backbone model for all methods.  %
Experiments with other backbone models are shown in \Cref{app:ablation_model}.
On images and videos, \gls{ntl} is built upon the penultimate layer output of a frozen ResNet-152 pre-trained on ImageNet.
\gls{ntl} is trained for one epoch, after which all $|\gQ|$ queries are labeled at once. %
The contamination ratio $\alpha$ in \gls{aloe} is estimated immediately after the querying step and then fixed for the remaining training process. 
We follow \citet{qiu2022latent} and set $\tilde{y}_i = 0.5$ for inferred anomalies. This accounts for the uncertainty of whether the sample truly is an anomaly. More details are given in \Cref{sec:implementation} and \Cref{alg:aloe}.

\begin{figure*}[t!]
    \centering
    \includegraphics[width=0.9\linewidth]{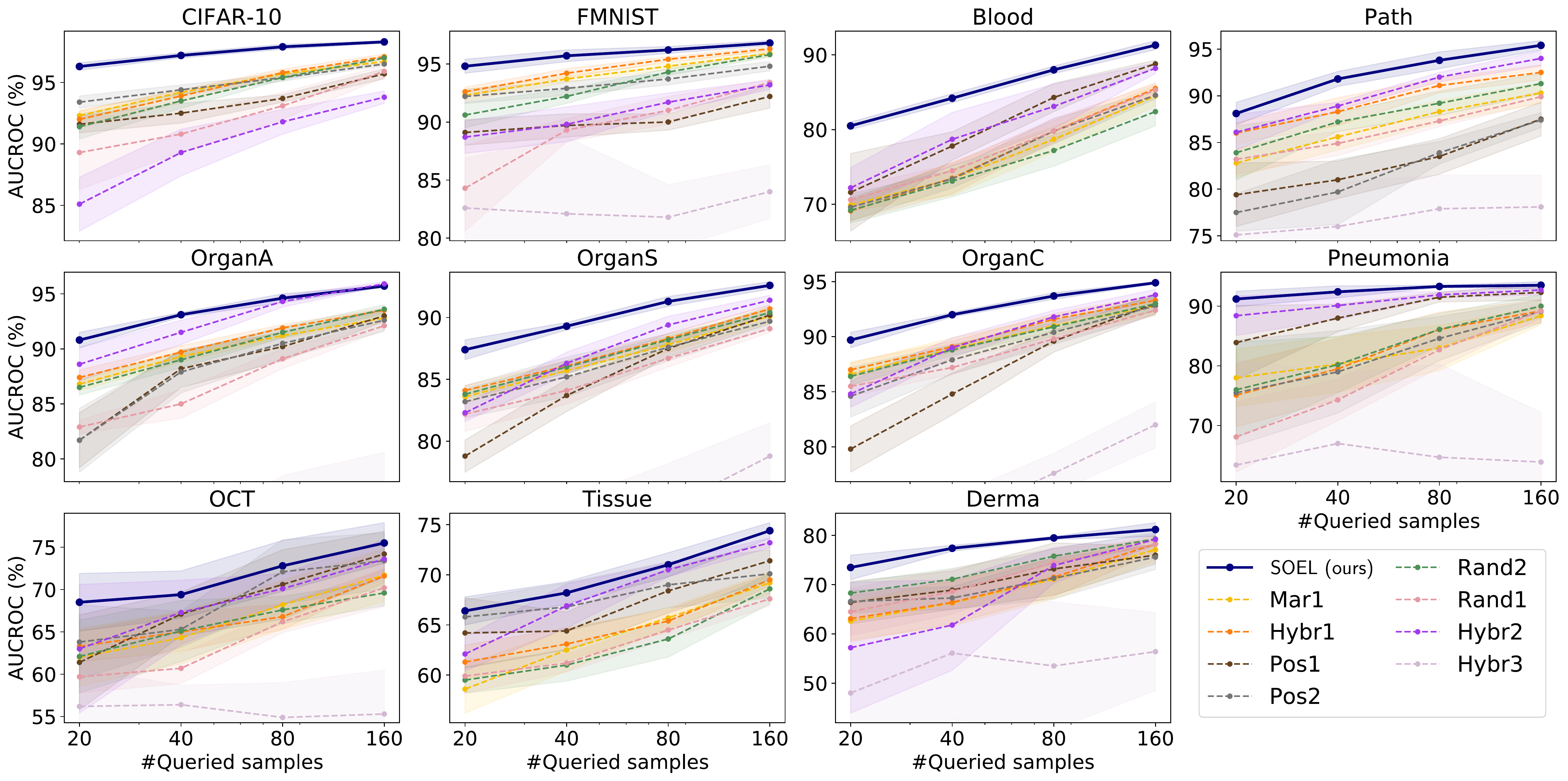}
    \vspace{-10pt}
    \caption{Running AUCs (\%) with different query budgets. Models are evaluated at $20, 40, 80, 160$ queries. \gls{aloe} performs the best among the compared methods on all query budgets.}
    \label{fig:img_data}
\vspace{-10pt}
\end{figure*}
\subsection{Experiments on Image Data}
\label{sec:image_exp}
We study \gls{aloe} on standard image benchmarks to establish how it compares to eight well-known baselines with various querying and training strategies.
Informative querying plays an important role in medical domains where expert labeling is expensive. Hence, we also study nine medical datasets from \citet{medmnistv2}. 
We describe the datasets, the evaluation protocol, and finally the results of our study.

\textbf{Image Benchmarks.}
We experiment with two popular image benchmarks: CIFAR-10 and Fashion-MNIST. These have been widely used in previous papers on %
deep \gls{ad}  
\citep{ruff2018deep,golan2018deep,hendrycks2019using,bergman2020classification}.

\textbf{Medical Images.}
Since medical imaging is an important practical application of \gls{ad}, we also study \gls{aloe} on medical images. The datasets we consider cover different data modalities (e.g., X-ray, CT, electron microscope) and their characteristic image features can be very different from natural images.   
Our empirical study includes all 2D image datasets presented in \citet{medmnistv2}
that have more than 500 samples in each class, including \rbt{Blood, OrganA, OrganC, OrganS, OCT, Pathology, Pneumonia, and Tissue.} 
We also include \rbt{Dermatoscope} but restricted to the classes with more than 500 training samples. %

\textbf{Evaluation Protocol.}
We follow the community standard known as the ``one-vs.-rest'' protocol to turn these classification datasets into a test-bed for \gls{ad} \citep{ruff2018deep,hendrycks2019using,bergman2020classification}.
While respecting the original train and test split of these datasets, the protocol iterates over the classes and treats each class in turn as normal. Random samples from the other classes are used to contaminate the data. The training set is then a mixture of unlabeled normal and abnormal samples with a contamination ratio of $10\%$ \citep{ruff2019deep,wang2019effective,qiu2022latent}. 
This protocol can simulate a ``human expert'' to provide labels for the queried samples because the datasets provide ground-truth class labels.
The reported results (in terms of AUC $\%$) for each dataset are averaged over the number of experiments (i.e., classes) and over five independent runs. %

\textbf{Results.}
We report the evaluation results of our method (\gls{aloe}) and the eight baselines on all eleven image datasets in \Cref{tab:img_results}. 
When querying 20 samples, our proposed method \gls{aloe} significantly outperforms the best-performing baseline by 6 percentage points on average across all datasets.
We also study detection performance as the query budget increases from $20$ to $160$ in \Cref{fig:img_data}. %
The results show that, with a small budget of 20 samples, \gls{aloe} (by querying diverse and informative samples) makes better usage of the labels than the other baselines and thus leads to better performance by a large margin. As more samples are queried, the performance of almost all methods increases but even for $160$ queries when the added benefit from adding more queries starts to saturate, \gls{aloe}  still outperforms the baselines.
\vspace{-0.5em}

\begin{table*}[t!]
\vspace{-5pt}
    \caption{F1-score ($\%$) with standard deviation for anomaly detection on tabular data when the query budget $|\gQ|=10$. \gls{aloe} performs the best on 3 of 4 datasets and outperforms all baselines by 3.2 percentage points on average.}%
\label{tab:tab_results}
    \footnotesize
    \centering
  \begin{tabular}{l|ccccccccc}
  \toprule
     & \bfseries\Margin & \bfseries\HybridA & \bfseries\PositiveB & \bfseries\PositiveA & \bfseries\RandomA & \bfseries\RandomB & \bfseries\HybridB & \bfseries\HybridC & \bfseries \gls{aloe} \\ 
  \midrule
\bfseries  BreastW  & 81.6$\pm$0.7	& 83.3$\pm$2.0	& 58.6$\pm$7.7	& 81.3$\pm$0.8	& 87.1$\pm$1.0	& 82.9$\pm$1.1	& 55.0$\pm$6.0	& 79.6$\pm$4.9	& \textbf{93.9$\pm$0.5}	\\ 
\bfseries Ionosphere & 91.9$\pm$0.3	& \textbf{92.3$\pm$0.5}	& 56.1$\pm$6.2	& 91.1$\pm$0.8	& 91.1$\pm$0.3	& 91.9$\pm$0.6	& 64.0$\pm$4.6	& 88.2$\pm$0.9	& 91.8$\pm$1.1	\\ 
 \bfseries Pima & 50.1$\pm$1.3	& 49.2$\pm$1.9	& 48.5$\pm$0.4	& 52.4$\pm$0.8	& 53.6$\pm$1.1	& 51.9$\pm$2.0	& 53.8$\pm$4.0	& 48.4$\pm$0.7	& \textbf{55.5$\pm$1.2}	\\ 
 \bfseries Satellite & 64.2$\pm$1.2	& 66.2$\pm$1.7	& 57.0$\pm$3.0	& 56.7$\pm$3.2	& 67.7$\pm$1.2	& 66.6$\pm$0.8	& 48.6$\pm$6.9	& 56.9$\pm$7.0	& \textbf{71.1$\pm$1.7}	\\ 
\midrule
\bfseries Average & 72.0 & 72.8 & 55.1 & 70.4 & 74.9 & 73.3 & 55.4 &  68.3 & \textbf{78.1} \\
  \bottomrule 
  \end{tabular}
 \vspace{-10pt}
\end{table*}
\subsection{Experiments on Tabular Data}
\label{sec:tab_exp}
Many practical use cases of \gls{ad} (e.g., in health care or cyber security) are concerned with tabular data.
For this reason, we study \gls{aloe} on four tabular datasets from various domains. We find that it outperforms existing baselines, even with as few as 10 queries. We also confirmed the fact that our deep models are competitive with classical methods for tabular data in \Cref{app:more-comp}.

\textbf{Tabular Datasets.} 
Our study includes the four multi-dimensional tabular datasets from the ODDS repository which have an outlier ratio of at least $30\%$. This is necessary to ensure that there are enough anomalies available to remove from the test set and add to the clean training set (which is randomly sub-sampled to half its size) to achieve a contamination ratio of $10\%$. The datasets are BreastW, Ionosphere, Pima, and Satellite. %
As in the image experiments, there is one round of querying, in which $10$ samples are labeled. For each dataset, we report the averaged F1-score (\%) with standard deviations over five runs with random train-test splits and random initialization. %

\textbf{Results.}
\gls{aloe} performs best on 3 of 4 datasets and outperforms all baselines by 3.2 percentage points on average. Diverse querying best utilizes the query budget to label the diverse and informative data points, yielding a consistent improvement over existing baselines on tabular data.

\subsection{Experiments on Video Data}
\label{sec:video_exp}
\begin{figure}[ht]
\vspace{-10pt}
    \centering
    \includegraphics[width=0.85\linewidth]{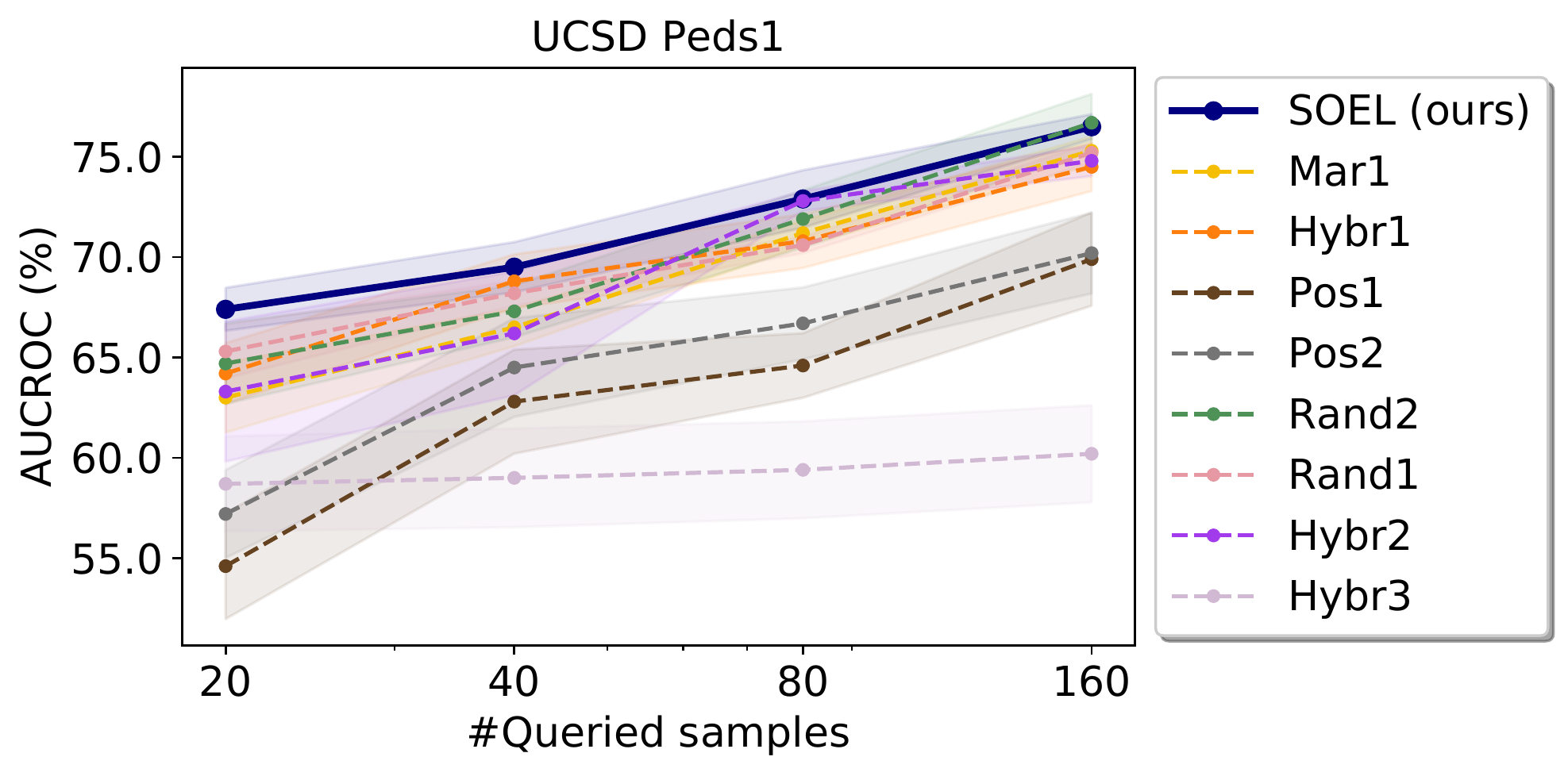}
\vspace{-10pt}
    \caption{Results on the video dataset UCSD Peds1 with different query budgets. \gls{aloe} achieves the leading performance.
    } 
    \label{fig:video_data}
 \vspace{-10pt}
\end{figure}

Detecting unusual objects in surveillance videos is an important application area for \gls{ad}. Due to the large variability in abnormal objects and suspicious behavior in surveillance videos, expert feedback is very valuable to train an anomaly detector in a semi-supervised manner. We use \gls{ntl} as the backbone model and  study \gls{aloe} on a public surveillance video dataset (UCSD Peds1). The goal is to detect abnormal video frames that contain non-pedestrian objects. %

Following \citet{pang2020self}, we subsample a mix of normal and abnormal frames for training (using an anomaly ratio of $0.3$) and use the remaining frames for testing.
Before running any of the methods, a ResNet pretrained on ImageNet is used to obtain a fixed feature vector for each frame. We vary the query budget from $|\gQ|=20$ to $|\gQ|=160$ and compare \gls{aloe} to all baselines. %
Results in terms of average AUC and standard error over five independent runs are reported in \Cref{fig:video_data}. \gls{aloe} consistently outperforms all baselines, especially for smaller querying budgets. %

\subsection{Additional Experiments}
\label{sec:add-exp}
In \Cref{app:ablation}, we provide additional experiments and ablations demonstrating \gls{aloe}'s strong performance and justifying modeling choices. The three most important findings are:
\vspace{-8pt}
\begin{itemize}[leftmargin=*,itemsep=-4pt]
\item {\bf \gls{aloe} vs. Active Learning:} Our framework is superior to its extension to the sequential active learning (\Cref{fig:img-data-seq}).
\item {\bf Varying Contamination Ratio:} \Cref{fig:robust-ratio} demonstrates that \gls{aloe} dominates under varying contamination ratios (1\%, 5\%, 20\%). In addition, \Cref{tab:alpha-estimate} confirms that \Cref{eq:alpha-is} reliably estimates 
 $\alpha$ on both CIFAR-10 and F-MNIST. 
\item {\bf Backbone Models:} \Cref{tab:backbone-dsvdd-mhrot} 
shows that \gls{aloe} also performs best for the backbone models  MHRot~\citep{hendrycks2019using} and DSVDD~\citep{ruff2018deep}.
\end{itemize}
\vspace{-8pt}
In addition, we provide an ablation of the temperature $\tau$ (\Cref{tab:ablation-tau}), a discussion on the effects of initialization randomness (\Cref{app:rand-init}), an ablation study of the pseudo-label $\tilde y$ values (\Cref{tab:ablation-pseudo-y}), a comparison to binary classification (\Cref{fig:img-data-bce}), an ablation of the \gls{aloe} loss components, an ablation of querying strategies (\Cref{fig:ablation-query}),
additional methods for inferring $y$  (\Cref{fig:semisuper-comparison}), and comparison to additional semi-supervised baselines (\Cref{fig:semisuper-comparison}, \Cref{tab:tab_knn}).

\section{Conclusion}
\label{sec:discussion}

We introduced semi-supervised outlier exposure with a limited labeling budget (\gls{aloe}). 
Inspired by a set of conditions that guarantee the generalization of anomaly score rankings from queried to unqueried data, we proposed to use a diversified querying strategy %
and a combination of two losses for queried and unqueried samples. By weighting the losses equally to each other and by estimating the unknown contamination rate from queried samples, we were able to make our approach free of its most important hyperparameters, making it easy to use. An extensive empirical study on images, tabular data, and video 
confirmed the efficacy of \gls{aloe} as a semi-supervised learning framework compatible with many existing losses for \gls{ad}.

\textbf{Limitations:} The success of our approach relies on several heuristics that we demonstrated were empirically effective but that cannot be proven rigorously. Estimation of the contamination ratio can be noisy when the query set is small---but the LOE loss is robust even under misspecification of the contamination ratio \citep{qiu2022latent}. The diversified sampling strategy becomes expensive when the dataset is large, but this can be mitigated by random data thinning.%

\textbf{Societal Impacts:} The use of human labels for anomaly detection runs the risk of introducing potential human biases in the definition of what is anomalous, particularly for datasets involving human subjects. Since our approach relies heavily on a relatively small number of human labels, the deployment of our approach with real human labelers would benefit by having guidelines for the labelers in terms of providing fair labels and avoiding amplification of bias.

\section*{Acknowledgements}
SM acknowledges support by the National Science Foundation (NSF) under an NSF CAREER Award, award numbers 2003237 and 2007719, by the Department of Energy under grant DE-SC0022331, by the HPI Research Center in Machine Learning and Data Science at UC Irvine, by the IARPA WRIVA program, and by gifts from Qualcomm and Disney. Part of this work was conducted within the DFG research unit FOR 5359 on Deep Learning on Sparse Chemical Process Data. MK acknowledges support by the Carl-Zeiss Foundation, the DFG awards KL 2698/2-1, KL 2698/5-1, KL 2698/6-1, and KL 2698/7-1, and the BMBF awards 03|B0770E and 01|S21010C.

The Bosch Group is carbon neutral. Administration, manufacturing and research activities do no longer leave a carbon footprint. This also includes GPU clusters on which the experiments have been performed.

\bibliography{refs}

\appendix
\onecolumn

\section{Theorem 1}
\label{app:thm1}

\begin{figure}[ht]
    \centering
    \includegraphics[width=0.45\linewidth]{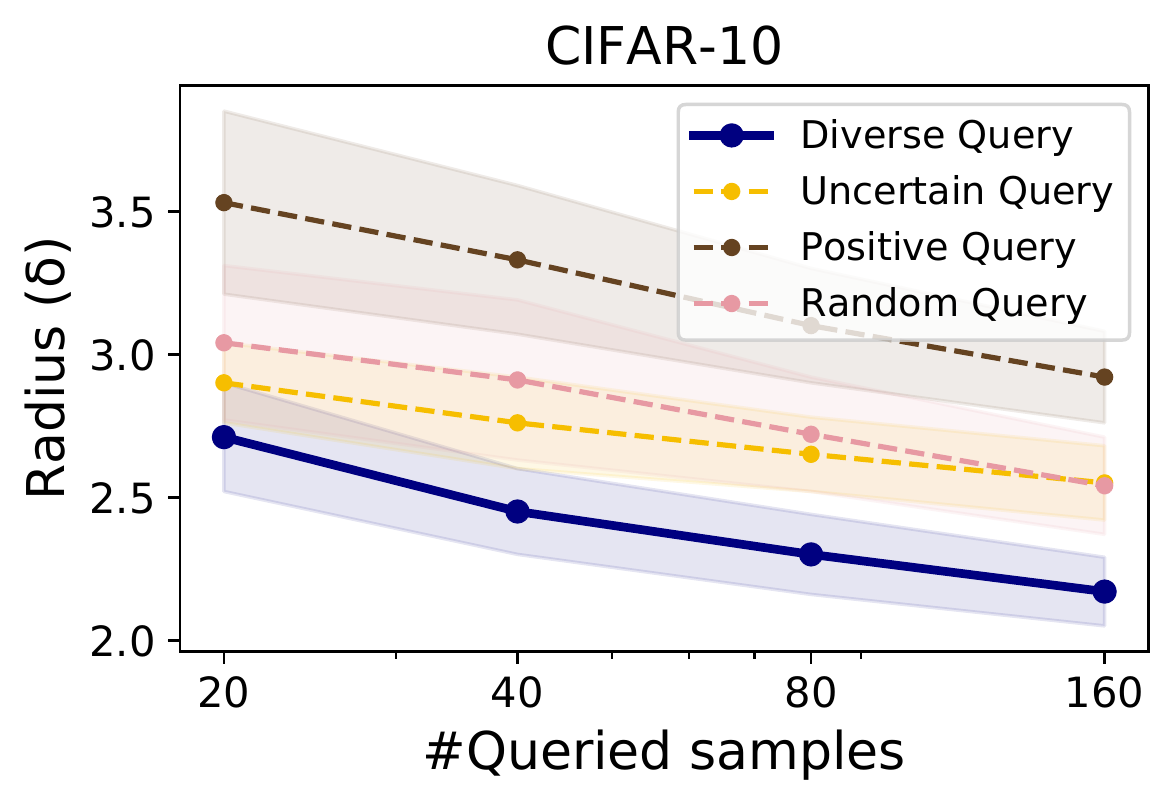}
    \includegraphics[width=0.45\linewidth]{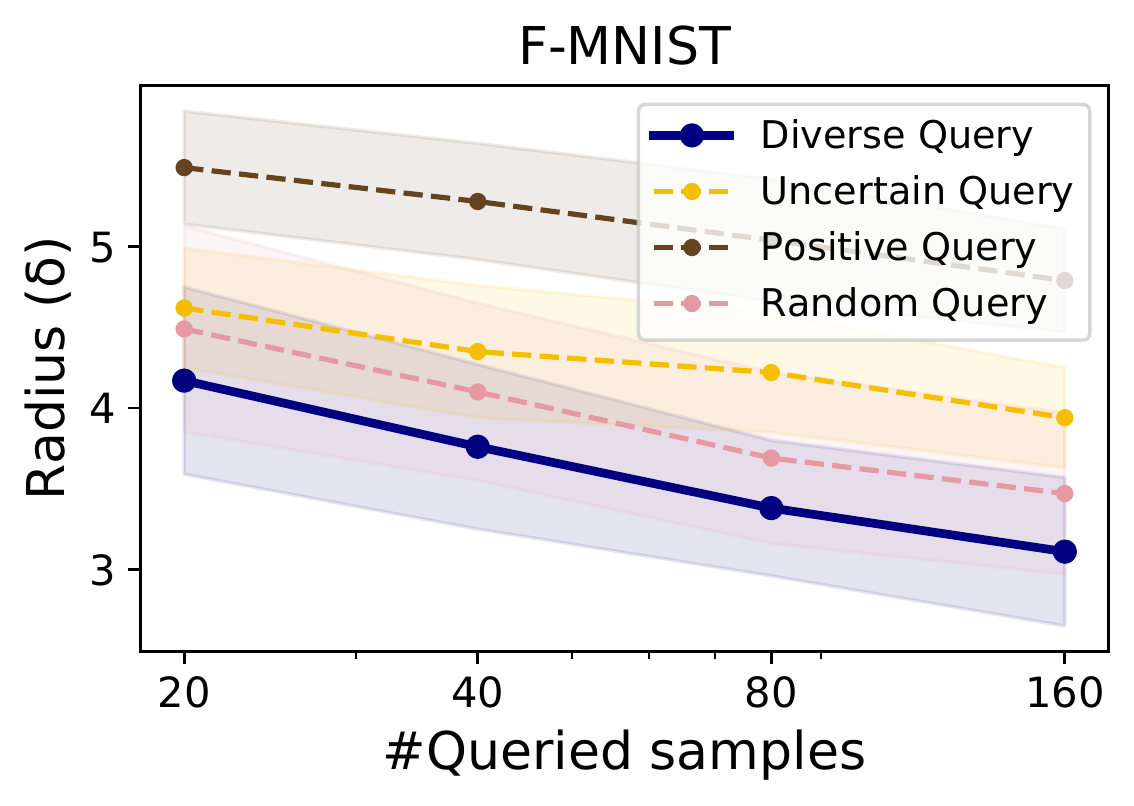}
    \caption{Cover radius $\delta$ (\Cref{eq:delta-est}) resulted from different querying strategies on the first class of CIFAR-10 and F-MNIST. Diverse queries systematically have smaller cover radius than other querying strategies.} 
    \label{fig:thm1-delta}
\end{figure}
\begin{figure}[ht]
    \centering
    \includegraphics[width=0.45\linewidth]{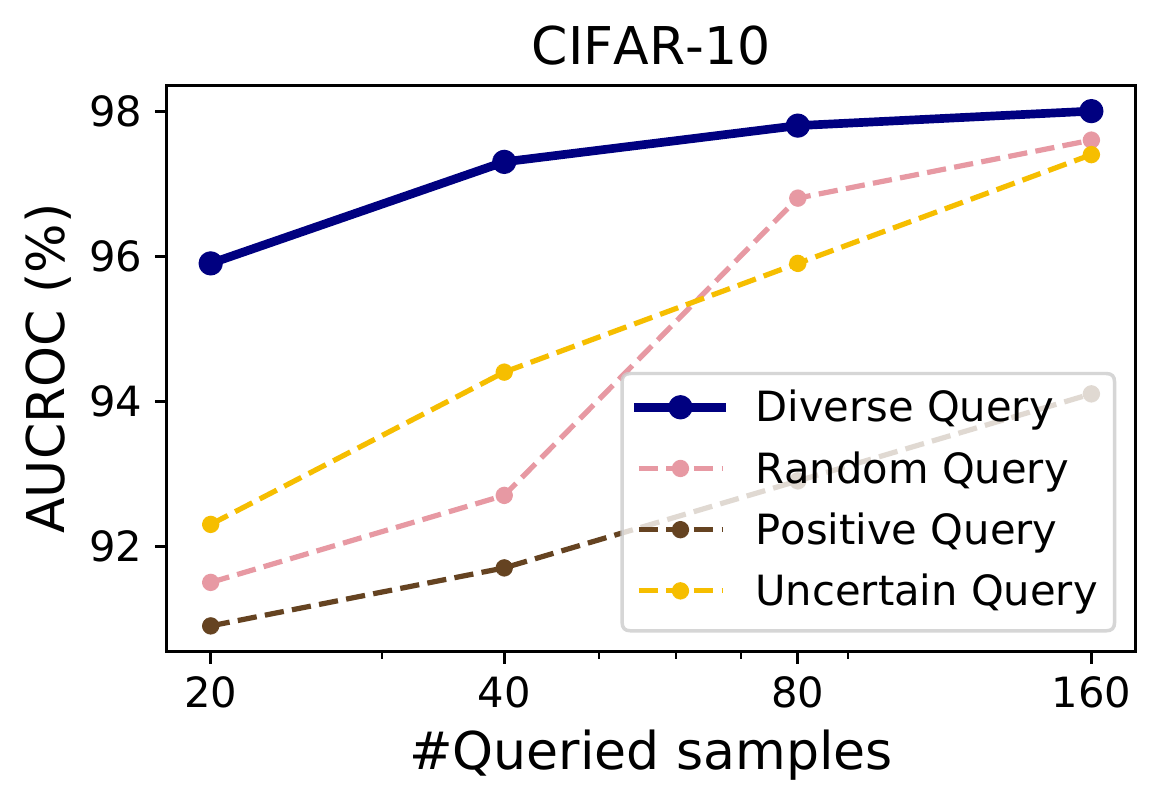}
    \includegraphics[width=0.45\linewidth]{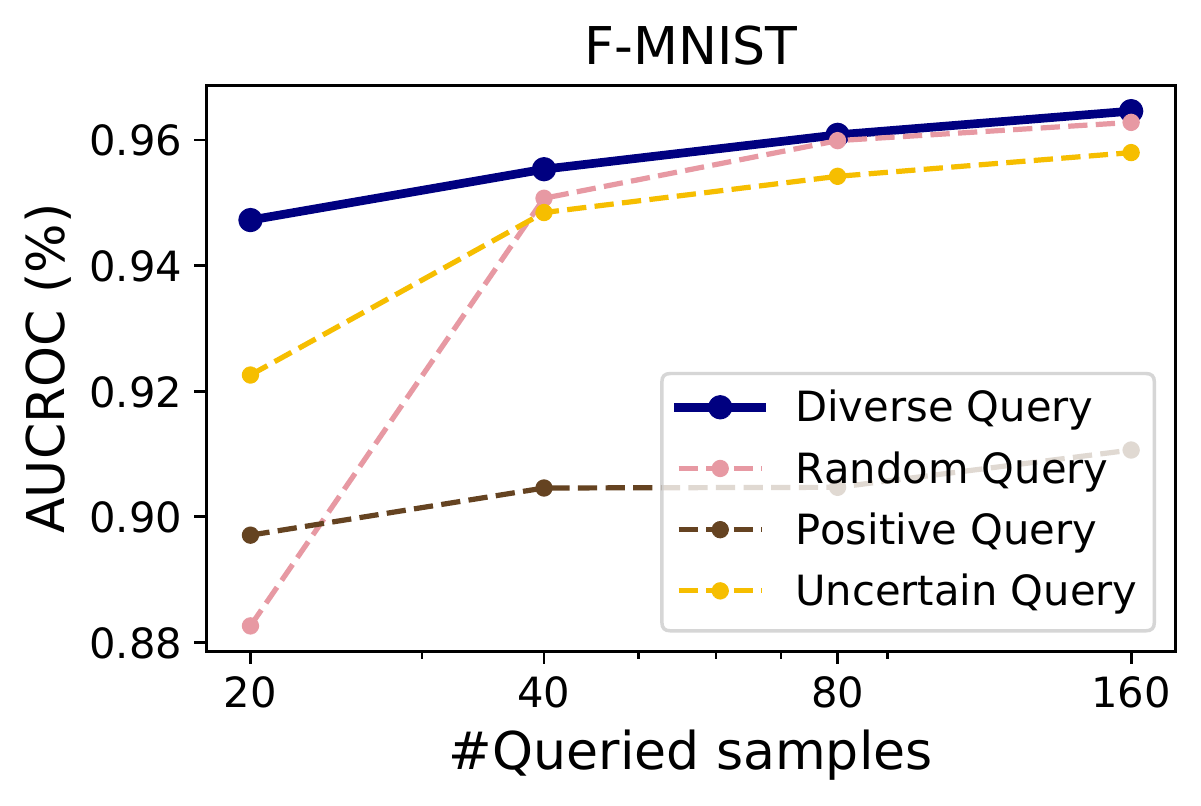}
    \caption{Ranking performance of unlabeled data. AUC of unqueried data is evaluated using the fitted anomaly detector on the queried data. Our proposed diverse querying (\kmeans) provides better ranking of the unlabeled data.} 
    \label{fig:ranking-unquery}
\end{figure}

\begin{proof}%
\vspace{-0.5em}
Since $\s$ is $\lambda_s$-Lipschitz continuous and $\vu_a$ and $\vu_n$ are assumed to be closer than $\delta$ to $\vx_a$ and $\vx_n$ respectively, we have 
$\s(\vx_a) - \delta\lambda_s\leq\s(\vu_a)$ and  $-\s(\vx_n) - \delta\lambda_s\leq-\s(\vu_n)$. Adding the inequalities and using the condition $\s(\vx_a)-\s(\vx_n) \geq 2\delta\lambda_s$, yields $0 \leq \s(\vx_a) - \s(\vx_n) - 2\delta\lambda_s\leq \s(\vu_a) - \s(\vu_n)$, which proves $\s(\vu_a) \geq \s(\vu_n)$. 
\vspace{-0.5em}
\end{proof}

In \Cref{thm:rank}, we considered using the fixed-radius neighborhood ($\delta$-ball) of the queried data as the cover of the whole dataset, and mentioned diverse querying has a smaller radius than other querying strategies. In this section, we will empirically verify this fact and further illustrate diverse querying leads to good ranking of un-queried data (see also \Cref{fig:ablation-query} on test data). 

As defined in \Cref{thm:rank}, the radius is the smallest distance that is required for any un-queried sample to be covered by the neighborhood of a queried sample of the same type. Mathematically, we compute the radius as 
\begin{equation}
\label{eq:delta-est}
    \delta = \max_{i\in\gU}\min_{j\in\gQ, y_i=y_j}d(\vx_i, \vx_j),
\end{equation}
where we adopt the euclidean distance in the feature space for a meaningful metric $d$. We apply NTL on the first class of CIFAR-10 and F-MNIST dataset. We make queries with different budgets, after which we compute $\delta$ by \Cref{eq:delta-est}. We repeat this procedure for 100 times and report the mean and standard deviation in  \Cref{fig:thm1-delta}. We compared four querying strategies: diverse queries (\kmeans), uncertain queries (\Margin), positive queries (\PositiveB), and random queries (\RandomA). It shows that diverse queries significantly lead the smallest radius $\delta$ among the compared strategies on all querying budgets.

Next, we provide an empirical, overall justification of \Cref{thm:rank} (see also \Cref{fig:ablation-query} on test data). An implication of \Cref{thm:rank} is that, assuming anomaly scores are fixed, a smaller $\delta$ will satisfy the large anomaly score margin ($\s(\vx_a)-\s(\vx_n)$) more easily, hence it is easier for $\s$ to correctly rank the remaining unlabeled points. To justify this implication, we need a metric of ranking. AUC satisfies this requirement as it is alternatively defined as~\citep[10.5.2]{mohri2018foundations}\citep{cortes2003auc}
\begin{equation*}
    \text{AUC}=\frac{1}{|\gU_0| + |\gU_1|}\sum_{n\in\gU_0, a\in\gU_1} \I(\s(\vu_a)>\s(\vu_n))\approx P_{n\in\gU_0, a\in\gU_1}(\s(\vu_a)>\s(\vu_n))
\end{equation*}
which measures the probability of ranking unlabeled samples $\vu_a$ higher than $\vu_n$ in terms of their scores. $\gU=\gU_0\bigcup \gU_1$ is the un-queried data indices and $\gU_0$ and $\gU_1$ are disjoint un-queried normal and abnormal data sets respectively. $\vu_a$ and $\vu_n$ are instances of each kind. We conducted experiments on CIFAR-10 and F-MNIST, where we trained an anomaly detector (NTL) on the queried data for 30 epochs and then compute the AUC on the remaining un-queried data. The results of four querying straties are reported in \Cref{fig:ranking-unquery}, which 
shows that our proposed diverse querying strategy generalizes the anomaly score ranking the best  to the unqueried data among the compared strategies, testifying our analysis in the main paper. A consequence is that diverse querying can provide accurate assignments of the latent anomaly labels, which will further help learn a high-quality of anomaly detector through the unsupervised loss term in \Cref{eqn:loss-2}.

\paragraph{Optimality of Cover Radius.} Although \kmeans greedily samples the queries which may have a sub-optimal cover radius, greedy sampling strategies for selecting a diverse set of datapoints in a multi-dimensional space are known to produce nearly optimal solutions \citep{krause2014submodular}, with significant runtime savings over more sophisticated search methods. As a results, we follow common practice (e.g. \citet{arthur2007k}) and also use the greedy approach. We check the diversity of the rustling query set by comparing all sampling strategies considered in the paper in terms of data coverage. Figure 4 shows that the greedy strategy we use achieves the best coverage, i.e. results in the most diverse query set.

\paragraph{On the Assumptions of \Cref{thm:rank}.} In the proof, we assume a Lipschitz continuous $S$ and a large margin between $S(\vx_a)$ and $S(\vx_n)$. Lipschitz continuity serves as a working assumption and is a common assumption when analyzing optimization landscapes of deep learning.  Lipschitz continuity can be controlled by the strength of regularization on the model parameters. 
The large margin condition is achieved by optimizing our loss function. The supervised anomaly detection loss encourages a large margin as it minimizes the anomaly score of queried normal data and maximizes the score of the queried abnormal data. If the anomaly score function doesn’t do well for the queried samples, then it should be optimized further. Our empirical results also show this is a reasonable condition.

\section{Theorem 2}
\label{sec:assum-veri}

In this section, we will empirically justify the assumptions we made in \Cref{sec:alpha} that are used to build an unbiased estimator of the anomaly ratio $\alpha$ (\Cref{eq:alpha-is}). We will also demonstrate the robustness of the estimation under varying $\alpha$. 

\subsection{Proof}
\begin{proof}
\vspace{-0.5em}
Let A1 and A2 denote Assumption 1 and 2, respectively. Furthermore, 
let $q(\vx_1, ..., \vx_{|\gQ|})$ and $q_s(s_1, ..., s_{|\gQ|})$ denote the query distribution in the data and anomaly score spaces, respectively. A2 assumes $\Ia_s(s):=\Ia_s(S(\vx))=\Ia(\vx)$ for all $\vx$. So the expectation of \Cref{eq:alpha-is} is 
\begin{align*}
&\E[\hat \alpha] = \E_{q(\vx_1, ..., \vx_{|Q|})}\left[\frac{1}{|\gQ|}\sum_{i=1}^{|\gQ|}\frac{p_s(S(\vx_i))}{q_s(S(\vx_i))}\Ia(\vx)\right]  
\stackrel{A2}{=} \E_{q_s(s_1, ..., s_{|Q|})}\left[\frac{1}{|\gQ|}\sum_{i=1}^{|\gQ|}\frac{p_s(s_i)}{q_s(s_i)}\Ia_s(s_i)\right] \\ 
& \stackrel{A1}{=} \E_{\prod_{i=1}^{|Q|}q_s(s_i)}\left[\frac{1}{|\gQ|}\sum_{i=1}^{|\gQ|}\frac{p_s(s_i)}{q_s(s_i)}\Ia_s(s_i)\right] 
= \frac{1}{|\gQ|}\sum_{i=1}^{|\gQ|} \E_{q_s(s_i)}\left[\frac{p_s(s_i)}{q_s(s_i)}\Ia_s(s_i)\right]
= \E_{p_s(s)}[\Ia_s(s)] \\
&= \E_{p(\vx)}[\Ia_s(S(\vx))]
\stackrel{A2}{=}\E_{p(\vx)}[\Ia(\vx)]=\alpha
\end{align*}
where the change of variables makes necessary assumptions, including the existence of density functions. \qedhere

\vspace{-0.5em}
\end{proof}

\subsection{Assumption 1}
\label{sec:assum-veri-1}
\begin{figure}[ht]
    \centering
    \includegraphics[width=0.4\linewidth]{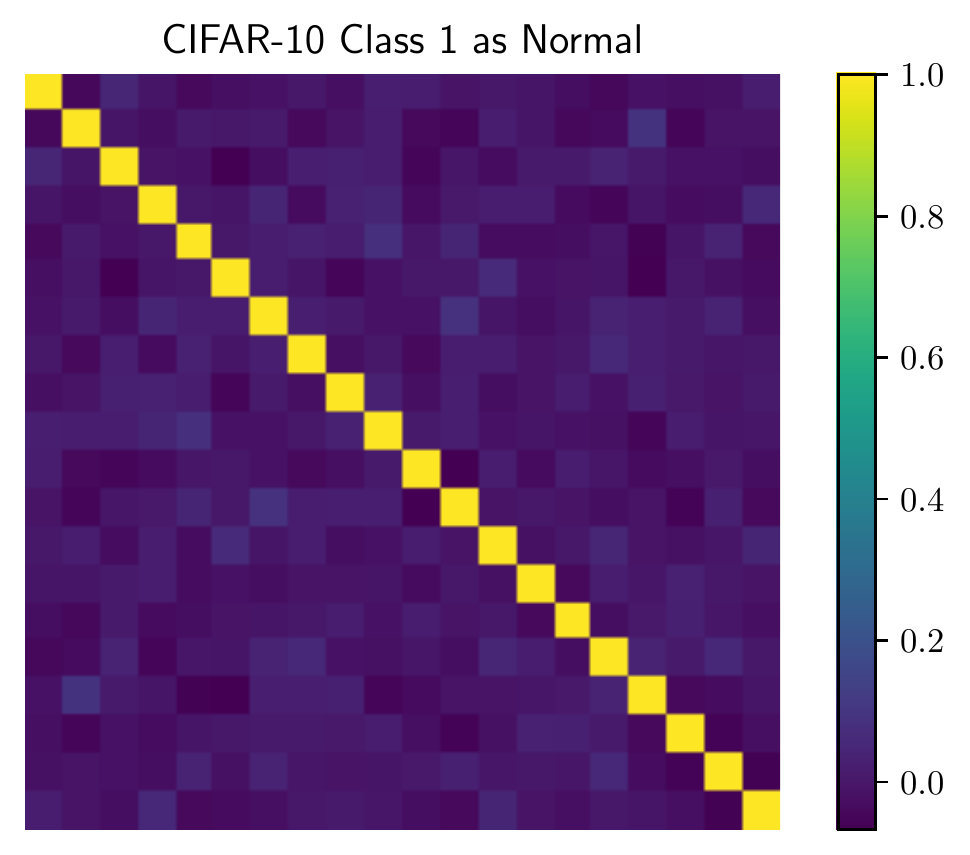}
    \includegraphics[width=0.4\linewidth]{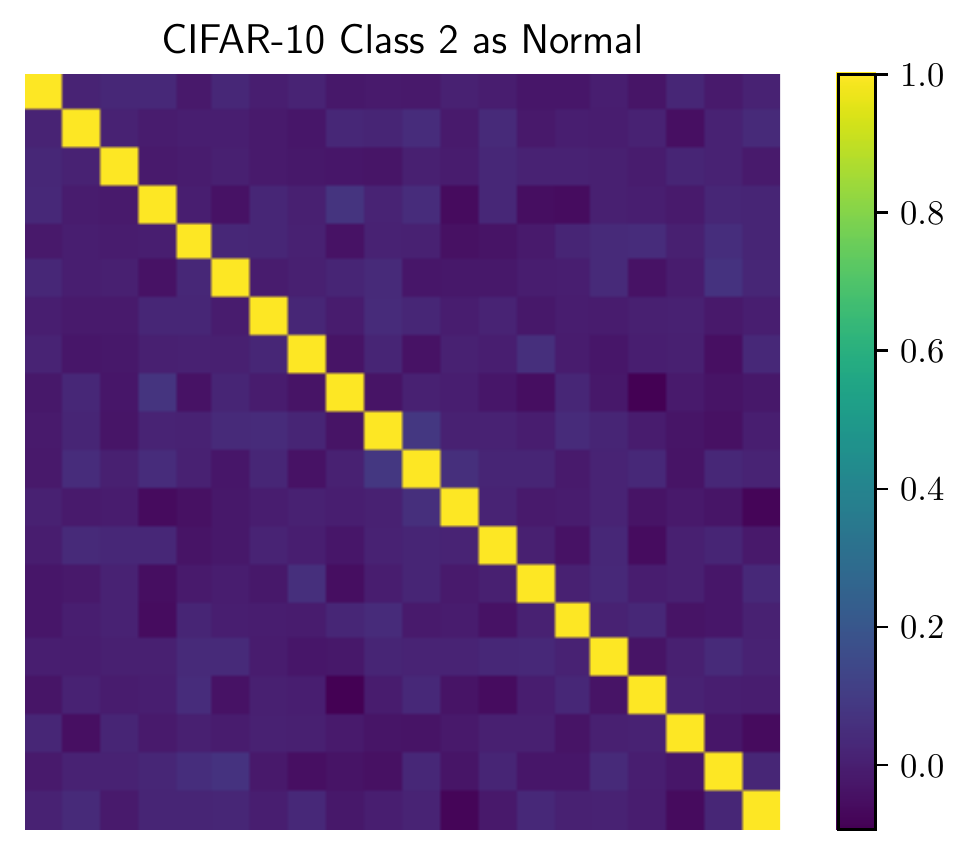}
    \includegraphics[width=0.4\linewidth]{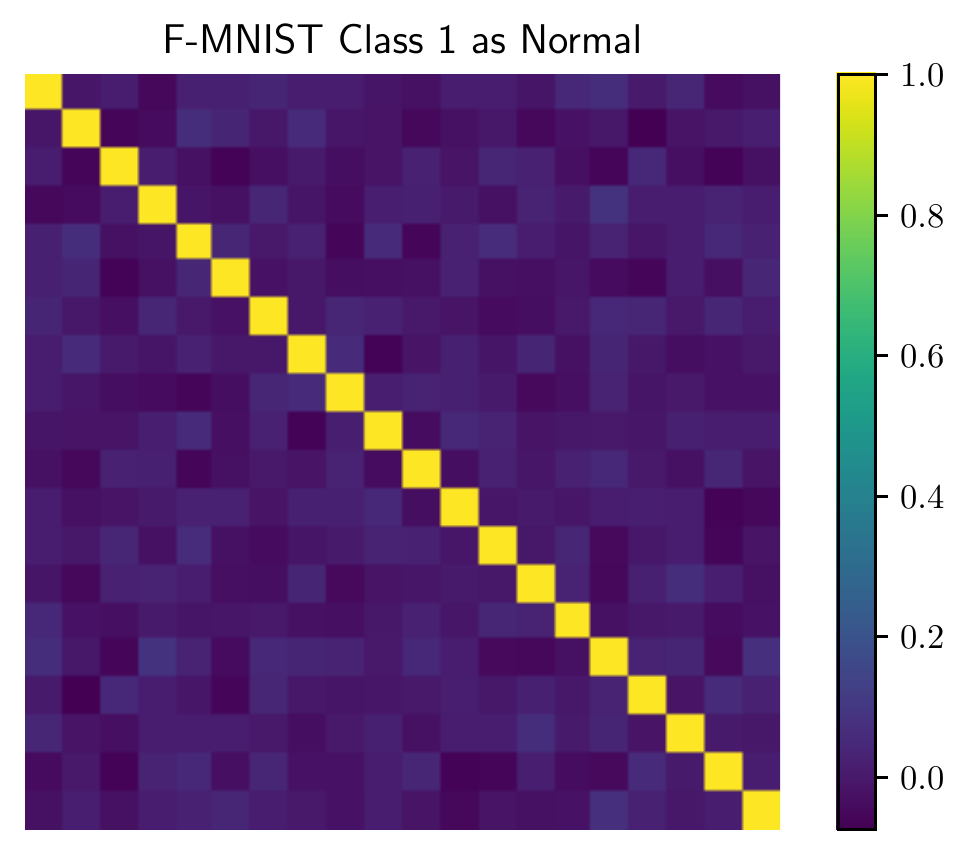}
    \includegraphics[width=0.4\linewidth]{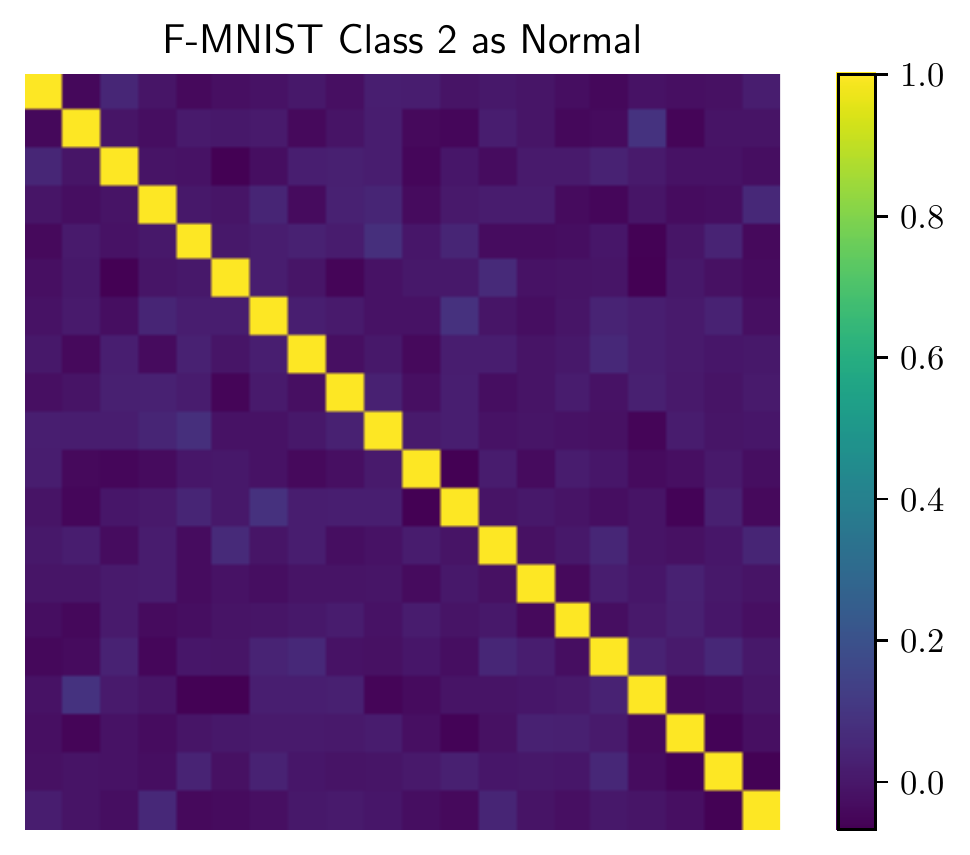}
    \caption{Anomaly score correlation matrix $\langle S(\vx_i),S(\vx_j)\rangle$, where $\vx_i$ and $\vx_j$ are jointly sampled in the same query set. The result indicates that anomaly scores can be considered as approximately independent random variables.
    } 
    \label{fig:corr-mat}
\end{figure}
We verify Assumption 1 by showing the correlation matrix in \Cref{fig:corr-mat}, where we jointly queried 20 points with diversified querying strategy and repeated 1000 times on two classes of CIFAR-10 and F-MNIST. Then the correlation between each pair of points are computed and placed in the off-diagonal entries. For each matrix, we show the average, maximum, and minimum of the off-diagonal terms
\begin{itemize}
    \item CIFAR-10 Class 1: -0.001, 0.103, -0.086
    \item CIFAR-10 Class 2: -0.001, 0.085, -0.094
    \item F-MNIST Class 1: -0.001, 0.081, -0.075
    \item F-MNIST Class 2: -0.005, 0.087, -0.067
\end{itemize}
Which shows the correlations $\langle S(\vx_i), S(\vx_j)\rangle$ are negligible, and the anomaly scores can be considered approximately independent random variables.

\subsection{Assumption 2}
\label{sec:assum-veri-3}
We verify Assumption 2 by counting the violations, i.e., $S(\vx_i)=S(\vx_j)$ but $y(\vx_i)\neq y(\vx_j)$ (because Assumption 2 states $\Ia_s(s_i)=\Ia(\vx_i)$ and $\Ia_s(s_j)=\Ia(\vx_j)$, $S(\vx_i)=S(\vx_j)$ implies $y(\vx_i)=\Ia_s(s_i)=\Ia_s(s_j)=y(\vx_j)$. The negation is $S(\vx_i)=S(\vx_j)$ and $y(\vx_i)\neq y(\vx_j)$.). We run the experiments on both CIFAR-10 and FMNIST. We apply the "one-vs.-rest" setup for both datasets and set the first class as normal and all the other classes as abnormal. We set the ground-truth anomaly ratio as 0.1. After the initial training, we count the pairs of data points that satisfy $S(\vx_i)=S(\vx_j)$ but $y(\vx_i)\neq y(\vx_j)$ for $i\neq j$. Our validation shows that on FMNIST, among 6666 training data points, there are 38 pairs of matching scores, and none of them have opposite labels, and on CIFAR-10, among 5555 training data points, the numbers are 21 and 3, respectively.

\subsection{Contamination Ratio Estimation}
\label{app:alpha-est}

\begin{table}[ht]
\vspace{-5pt}
	\caption{Estimated contamination ratios on CIFAR-10 and F-MNIST when $|\gQ|=40$ and the backbone model is NTL. The first row shows the true contamination ratio ranging from $1\%$ to $45\%$. The estimations are repeated 50 times.} %
	\label{tab:alpha-estimate}
	\small
	\centering
	\vspace{1pt}
	\begin{tabular}{lccccc}
        \toprule
		     & 1\% & 5\% & 10\% & 15\% &20\% \\
		\midrule
		CIFAR-10 &$0.5\% \pm 1.2\%$ & $6.0\% \pm 3.3\%$ & $12.0\% \pm 4.4\%$ & $15.3\% \pm 4.5\%$ 
        & 18.9\% $\pm$ 5.4\%\\
		F-MNIST &$1.0\% \pm 1.5\%$ & $3.8\% \pm 2.3\%$ &  $8.7\% \pm 4.1\%$ & $12.8\% \pm 5.3\%$ 
        & 19.3\% $\pm$ 5.1\%\\
        \bottomrule
	\end{tabular}

        \begin{tabular}{lccccc}
        \toprule
		     &25\% &30\% &35\% &40\% &45\% \\
		\midrule
		CIFAR-10 &26.2\% $\pm$ 6.0\%   & 30.6\% $\pm$ 5.5\% &35.8\% $\pm$ 6.9\%  &42.0\% $\pm$ 7.7\%  &47.2\% $\pm$ 6.7\%\\
		F-MNIST &27.9\% $\pm$ 6.4\%  &31.8\% $\pm$ 6.1\% &38.3\% $\pm$ 6.5\% &43.1\% $\pm$ 5.7\% &48.9\% $\pm$ 5.6\% \\
        \bottomrule
	\end{tabular}
\end{table}

We estimate the contamination ratio by \Cref{eq:alpha-is} under varying true ratios. This part shows the estimated contamination ratio when the query budget is $|\gQ|=40$. 
The estimations from the backbone model NTL is shown in~\Cref{tab:alpha-estimate}.
The first row contains the ground truth contamination rate, and the second and third row indicate the inferred values for two datasets, using our approach. Most estimates are withing the error bars and hence accurate. 
The estimation errors for low ground-truth contamination ratios are acceptable as confirmed by the sensitivity study in \citep{qiu2022latent} which concludes that the LOE approach still works well if the anomaly ratio is mis-specified within 5 percentage points. Interestingly, we find the estimation error increases somewhat with the contamination ratio. However, a contamination ratio larger than 40\% is rare in practice (most datasets should be fairly clean and would otherwise require additional preprocessing). In an anomaly detection benchmark (\url{https://github.com/Minqi824/ADBench}), none of the datasets have an anomaly ratio larger than 40\%.

\section{Baselines Details}
\label{sec:baselines}
In this section, we describe the details of the baselines in \Cref{tab:baselines} in the main paper. For each baseline method, we explain their query strategies and post-query training strategies we implement in our experiment. Please also refer to our codebase for practical implementation details.

\begin{itemize}[leftmargin=*,itemsep=2pt]
    \item \textbf{\RandomA.} This strategy used by \citet{ruff2019deep}  selects queries by sampling uniformly without replacement across the training set, resulting in the queried index set $\gQ=\{i_q \sim \text{Unif}(1,\cdots,N) | 1\leq q\leq |\gQ|\}$. After the querying, models are trained with a supervised loss function based on outlier exposure on the labeled data and with a one-class classification loss function on the unlabeled data,
    \begin{align}
    \label{eqn:randa}
        L_{\RandomA}(\theta) =  \frac{1}{|\gQ|}\sum_{j\in \gQ} \big(y_j \La(\vx_j) + (1-y_j) \Ln(\vx_j) \big) + \frac{1}{|\gU|}\sum_{i\in \gU} \Ln(\vx_i).
    \end{align}
    As in \gls{aloe} both loss contributions are weighted equally. $L_{\RandomA}(\theta)$ is minimized with respect to the backbone model parameters $\theta$.
    
    \item \textbf{\RandomB.} The querying strategy of \citet{trittenbach2021overview} samples uniformly among the top $50\%$ data ranked by anomaly scores without replacement. This leades to a random set of ``positive'' queries. After the queries are labeled, the training loss function is the same as $L_{\RandomA}(\theta)$ (\Cref{eqn:randa}).

    \item \textbf{\Margin.} After training the backbone model for one epoch, this querying strategy by ~\citet{gornitz2013toward} uses the $\alpha$-quantile ($s_\alpha$) of the training data anomaly scores to define a ``normality region''. Then the $|\gQ|$ samples closest to the margin $s_\alpha$ are selected to be queried. After the queries are labeled, the training loss function is the same as $L_{\RandomA}(\theta)$ (\Cref{eqn:randa}).
        Note that in practice we don't know the true anomaly ratio for the $\alpha$-quantile. In all experiment, we provide this querying strategy with the true contamination ratio of the dataset. Even with the true ratio, the ``\Margin'' strategy is still outperformed by \gls{aloe}.
    
    \item \textbf{\HybridA.} This hybrid strategy, also used by \citep{gornitz2013toward} combines the ``\Margin'' query with neighborhood-based diversification. %
    The neighborhood-based strategy selects samples with fewer neighbors covered by the queried set to ensure the samples' diversity in the feature space.
    We start by selecting the data index $\argmin_{1\leq i \leq N} \|s_i - s_\alpha\|$ into $\gQ$. Then the samples are selected sequentially without replacement by the criterion 
    \begin{align*}
        \argmin_{1\leq i \leq N} 0.5+\frac{|\{j\in \text{NN}_k(\phi(\vx_i)): j\in \gQ \}|}{2k} + \beta\frac{\|s_i - s_\alpha\|-\min_i\|s_i - s_\alpha\|}{\max_i\|s_i - s_\alpha\|-\min_i\|s_i - s_\alpha\|}
    \end{align*}
    where the inter-sample distance is measured in the feature space and the number of nearest neighbors is $k = \lceil N/|\gQ| \rceil$. We set $\beta=1$ for equal contribution of both terms. %
   After the queries are labeled, the training loss function is the same as $L_{\RandomA}(\theta)$ (\Cref{eqn:randa}).
    
    \item \textbf{\PositiveB.} This querying strategy by~\citet{pimentel2020deep} always selects the top-ranked samples ordered by their anomaly scores, $\argmax_{1\leq i\leq N} s_i$. 
    After the queries are labeled, the training loss only involves the labeled data
    \begin{align*}
        L_{\PositiveB}(\theta) = \frac{1}{|\gQ|}\sum_{j\in \gQ} \big(y_j \La(\vx_j) + (1-y_j) \Ln(\vx_j) \big).
    \end{align*}
    \citet{pimentel2020deep} use the logistic loss but we use the supervised outlier exposure loss. The supervised outlier exposure loss is shown to be better than the logistic loss in learning anomaly detection models~\citep{hendrycks2018deep,ruff2019deep}.
    
    \item \textbf{\PositiveA.} This approach of~\citep{barnabe2015active} uses the same querying strategy as \PositiveB, but the training is different. \PositiveA~also uses the unlabeled data during training. After the queries are labeled, the training loss function is the same as $L_{\RandomA}(\theta)$ (\Cref{eqn:randa}).

    \item \textbf{\HybridB.} This hybrid strategy by ~\citet{das2019active} makes positive diverse queries. It combines querying according to anomaly scores with distance-based diversification. \HybridB~selects the initial query $\argmax_{1\leq i \leq N} s_i$ into $\gQ$. Then the samples are selected sequentially without replacement by the criterion 
    \begin{align*}
        \argmax_{1\leq i \leq N} \frac{s_i-\min_is_i}{\max_is_i-\min_is_i} + \beta \min_{j\in\gQ} \frac{d(\vx_i, \vx_j)-\min_{a\neq b}d(\vx_a, \vx_b)}{\max_{a\neq b}d(\vx_a, \vx_b)-\min_{a\neq b}d(\vx_a, \vx_b)} 
    \end{align*}
    where $d(\vx_i, \vx_j)=  ||\phi(\vx_i) - \phi(\vx_j)||_2$.
    We set $\beta=1$ for equal contribution of both terms.
    After the queries are labeled, \citet{das2019active} use the labeled set to learn a set of weights for the components of an ensemble of detectors. For a fair comparison of active learning strategies, we use the labeled set to update an individual anomaly detector with parameters $\theta$ by optimizing the loss
    \begin{align*}
        L_{\HybridB}(\theta) = \frac{1}{|\gQ|}\sum_{j\in \gQ} \big(y_j \La(\vx_j) + (1-y_j) \Ln(\vx_j) \big).
    \end{align*}
    
    \item \textbf{\HybridC.} This baseline by ~\citep{ning2022deep} uses the same query strategy as \HybridB, but differs in the training loss function,
    \begin{align*}
        L_{\HybridC}(\theta) = \frac{1}{|\gQ|+|\gU|}\sum_{j\in \gQ} w_j(1-y_j) \Ln(\vx_j) + \frac{1}{|\gQ|+|\gU|}\sum_{i\in \gU} \hat w_i\Ln(\vx_i),
    \end{align*}
    where $w_j=2\sigma(d_j)$ and $\hat w_i=2 - 2\sigma(d_i)$ where $\sigma(\cdot)$ is the Sigmoid function and $d_i=10c_d\big(||\phi(\vx_i) - \vc_0||_2 - ||\phi(\vx_i) - \vc_1||_2 \big)$ where $\vc_0$ is the center of the queried normal samples and $\vc_1$ is the center of the queried abnormal samples in the feature space, and $c_d$ is the min-max normalization factor.
    
    We make three observations for the loss function. First, $L_{\HybridC}(\theta)$ filters out all labeled anomalies in the supervised learning part and puts a large weight (but only as large as 2 at most) to the true normal data that has a high anomaly score. Second, in the unlabeled data, $L_{\HybridC}(\theta)$ puts smaller weight (less than 1) to the seemingly abnormal data. Third, overall, the weight of the labeled data is similar to the weight of the unlabeled data. This is unlike \gls{aloe}, which weighs labeled data $|\gU|/|\gQ|$ times higher than unlabeled data.
\end{itemize}

\section{Implementation Details}
\label{sec:implementation}
In this section, we present the implementation details in the experiments. They include an overall description of the experimental procedure for all datasets, model architecture, data split, and details about the optimization algorithm.

\subsection{Experimental Procedure}
We apply the same  experimental procedure for each dataset and each compared method. The experiment starts with an unlabeled, contaminated training dataset with index set $\gU$. We first train the anomaly detector on $\gU$ for one epoch as if all data were normal. Then we conduct the diverse active queries at once and estimate the contamination ratio $\alpha$ by the importance sampling estimator \Cref{eq:alpha-is}. Lastly, we optimize the post-query training losses until convergence. The obtained anomaly detectors are evaluated on a held-out test set. The training procedure of \gls{aloe} is shown in \Cref{alg:aloe}.

\begin{algorithm}[t]
\caption{Training Procedure of \gls{aloe}}
\label{alg:aloe}
\textbf{Input}: Unlabeled training dataset $\mathcal{D}$,
querying budget $K$ \\
\textbf{Procedure}: \\
Train the model on $\gD$ for one epoch as if all data were normal;\\
Query $K$ data points from $\gD$ diversely resulting in a labeled set $\gQ$ and an unlabeled set $\gU$;\\
Estimate the contamination ratio $\alpha$ based on $\gQ$;\\
Finally train the model with $\{\gQ, \gU\}$ until convergence:\\
  For each iteration: \\
We construct a mini-batch with $\gQ$ and a subsampled mini-batch of $\gU$\\
The sample in $\gQ$ is up-weighted with $1/|\gQ|$ and the sample in $\gU$ is down-weighted with weight $1/|\gU|$\\
The training strategy for $\gQ$ is supervised learning; the training strategy for $\gU$ is LOE with the estimated anomaly ratio $\alpha$.

\end{algorithm}

\subsection{Data Split}
\paragraph{Image Data.} For the image data including both natural (CIFAR-10~\citep{krizhevsky2009learning} and F-MNIST~\citep{xiao2017/online}) and medical (MedMNIST~\citep{medmnistv2}) images, we use the original training, validation (if any), and test split. When contaminating the training data of one class, we randomly sample images from other classes' training data and leave the validation and test set untouched. Specifically for DermaMNIST in MedMNIST, we only consider the classes that have more than 500 images in the training data as normal data candidates, which include benign keratosis-like lesions, melanoma, and melanocytic nevi. We view all other classes as abnormal data. Different experiment runs have different randomness.

\paragraph{Tabular Data.} Our study includes the four multi-dimensional tabular datasets from the ODDS repository\footnote{\url{http://odds.cs.stonybrook.edu/}} which have an outlier ratio of at least $30\%$. . To form the training and test set for tabular data, we first split the data into normal and abnormal categories. We randomly sub-sample half the normal data as the training data and treat the other half as the test data. To contaminate training data, we randomly sub-sample the abnormal data into the training set to reach the desired $10\%$ contamination ratio; the remaining abnormal data goes into the test set. Different experiment runs have different randomness.

\paragraph{Video Data.} We use UCSD Peds1\footnote{\url{http://www.svcl.ucsd.edu/projects/anomaly/dataset.htm}}, a benchmark dataset for video anomaly detection. UCSD Peds1 contains 70 surveillance video clips --  34 training clips and 36 testing clips. Each frame is labeled to be abnormal if it has non-pedestrian objects and labeled normal otherwise. Making the same assumption as \citep{pang2020self}, we treat each frame independent and mix the original training and testing clips together. This results in a dataset of 9955 normal frames and 4045 abnormal frames. We then randomly sub-sample 6800 frames out of the normal frames  and 2914 frames out of the abnormal frames without replacement to form a contaminated training dataset with $30\%$ anomaly ratio. A same ratio is also used in the literature~\citep{pang2020self} that uses this dataset. The remaining data after sampling is used for the testing set, whose about $30\%$ data is  anomalous. Like the other data types, different experiment runs have different randomness for the training dataset construction.

\subsection{Model Architecture}
The experiments involve two anomaly detectors, \gls{ntl} and \gls{mhrot}, and three data types. 

\paragraph{NTL on Image Data and Video Data.} For all images (either natural or medical) and video frames, we extract their features by feeding them into a ResNet152 pre-trained on ImageNet and taking the penultimate layer output for our usage. The features are kept fixed during training. We then train an \gls{ntl} on those features. We apply the same number of transformations, network components, and anomaly loss function $\La(\vx)$, as when \citet{qiu2022latent} apply \gls{ntl} on the image data.%

\paragraph{NTL on Tabular Data.} We directly use the tabular data as the input of \gls{ntl}. We apply the same number of transformations, network components, and anomaly loss function $\La(\vx)$, as when \citet{qiu2022latent} apply \gls{ntl} on the tabular data.%

\paragraph{MHRot on Image Data.} We use the raw images as input for \gls{mhrot}. We set the same transformations, \gls{mhrot} architecture, and anomaly loss function as when \citet{qiu2022latent} apply \gls{mhrot} on the image data. %

\paragraph{DSVDD on Image Data.} For all images (either natural or medical), we build DSVDD on the features from the penultimate layer of a ResNet152 pre-trained on ImageNet. The features are kept fixed during training. The neural network of DSVDD is a three-layer MLP with intermediate batch normalization layers and ReLU activation. The hidden sizes are $[1024,512,128]$.

\subsection{Optimization Algorithm}
\label{app:optim}
\begin{table}[ht]
    \centering
    \small
    \begin{tabular}{ll|cccc}
    \toprule
    Model & Dataset &  Learning Rate & Epoch & Minibatch Size & $\tau$\\
    \midrule
    \multirow{5}{*}{NTL} & CIFAR-10 &  1e-4 & 30 & 512 & 1e-2 \\
        & F-MNIST &  1e-4 & 30 & 512 & 1e-2 \\
        & MedMNIST & 1e-4 & 30 & 512 & 1e-2 \\
        & ODDS & 1e-3 & 100 & $\lceil N/5 \rceil$ & 1e-2 \\
        & UCSD Peds1 & 1e-4 & 3$^*$ & 192 & 1e-2 \\
    \midrule
    \multirow{3}{*}{MHRot} & CIFAR-10 &  1e-3 & 15 & 10 & N/A \\
        & F-MNIST &  1e-4 & 15$^{**}$ & 10 & N/A \\
        & MedMNIST &  1e-4 & 15 & 10 & N/A \\
    \midrule
    \multirow{3}{*}{Deep SVDD} & CIFAR-10 & 1e-4 & 30 & 512 & 1e-2  \\
        & F-MNIST  & 1e-4 & 30 & 512 & 1e-2\\
        & MedMNIST & 1e-4 & 30 & 512 & 1e-2 \\
    \bottomrule
    \multicolumn{5}{l}{{\small $^*$\HybridB, \HybridC, \PositiveB, and \PositiveA~train 30 epochs. All other methods train 3 epochs.}}\\
    \multicolumn{5}{l}{{\small $^{**}$\gls{aloe} train 3 epochs. }}
    \end{tabular}
    \caption{A summary of optimization parameters for all methods.}
    \label{tab:opt-param}
\end{table}
In the experiments, we use Adam~\citep{kingma2014adam} to optimize the objective function to find the local optimal anomaly scorer parameters $\theta$. For Adam, we set $\beta_1=0.9, \beta_2=0.999$ and no weight decay for all experiments. 

To set the learning rate, training epochs, minibatch size for MedMNIST, we find the best performing hyperparameters by evaluating the method on the validation dataset. We use the same hyperparameters on other image data. For video data and tabular data, the optimization hyperparameters are set as recommended by ~\citet{qiu2022latent}. In order to choose $\tau$ (in \Cref{eq:query-prob}), we constructed a validation dataset of CIFAR-10 to select the parameter $\tau$ among \{1, 1e-1, 1e-2, 1e-3\} and applied the validated $\tau$ (1e-2) on all the other datasets in our experiments. Specifically, we split the original CIFAR-10 training data into a training set and a validation set. After validation, we train the model on the original training set again. We summarize all optimization hyperparameters in \Cref{tab:opt-param}.

When training models with \gls{aloe}, we resort to the block coordinate descent scheme that update the model parameters $\theta$ and  the pseudo labels $\tilde{\vy}$ of unlabeled data in turn. In particular, we take the following two update steps iteratively:
\begin{itemize}
    \item update $\theta$ by optimizing \Cref{eqn:loss-2} given $\tilde{\vy}$ fixed;
    \item update $\tilde{\vy}$ by sovling the constrained optimization in \Cref{sec:aloe-loss} given $\theta$ fixed;
\end{itemize}

Upon updating $\tilde{\vy}$, we use the LOE$_S$ variant~\citep{qiu2022latent} for the unlabeled data. We set the pseudo labels $\tilde\vy$ by performing the optimization below
\begin{align*}
      \min_{\tilde{\vy} \in \{0,0.5\}^{|\mathcal{U}|}} &~ \frac{1}{|\mathcal{U}|}\sum_{i \in\mathcal{U}} \tilde{y}_i\La(\vx_i) + (1-\tilde{y}_i)\Ln(\vx_i)  \qquad
      \text{s.t.} &~ \sum_{i=1}^{|\mathcal{U}|} \tilde{y}_i = \frac{\tilde\alpha|\mathcal{U}|}{2},
\end{align*}
where $\tilde\alpha$ is the updated contamination ratio of $\gU$ after the querying round, $\tilde \alpha = \big(\alpha N - \sum_{j\in\gQ} \Ia(\vx_j)\big) / |\gU|$, and $\alpha$ is computed by \Cref{eq:alpha-is} given $\gQ$. The solution is to rank the data by $\Ln(\vx)-\La(\vx)$ and label the top $\tilde\alpha$ data abnormal (equivalently setting $\tilde y = 0.5$) and all the other data normal (equivalently $\tilde y = 0$).

When we compute the Euclidean distance in the feature space, we construct the feature vector of a sample by concatenating all its encoder representations of different transformations. For example, if the encoder representation has 500 dimensions and the model has 10 transformations, then the final feature representation has $10\times 500 = 5000$ dimensions.

\rbt{
\subsection{Time Complexity}
\label{app:time-complexity}
Regarding the time complexity, the optimization uses stochastic gradient descent. The complexity of our querying strategy is $O(KN)$ where $K$ is the number of queries and $N$ is the size of the training data. This complexity can be further reduced to $O(K\log N)$ with a scalable extension of \kmeans~\citep{bahmani2012scalable}.
}

\section{Additional Experiments and Ablation Study}
\label{app:ablation}
The goal of this ablation study is to show the generality of \gls{aloe}, to better understand the success of \gls{aloe}, and to disentangle the benefits of the training objective and the querying strategy. To this end, we applied \gls{aloe} to different backbone models and different data forms (raw input and embedding input),  performed specialized experiments to compare the querying strategies, to demonstrate the optimality of the proposed weighting scheme in \Cref{eqn:loss-2}, and to validate the detection performance of the estimated ratio by \Cref{eq:alpha-is}. We also compared \gls{aloe} against additional baselines including semi-supervised learning frameworks and shallow anomaly detectors.

\subsection{Randomness of Initialization}
\label{app:rand-init}
Random Initialization affects both the queried samples and downstream performance. To evaluate the effects, we ran all experiments 5 times with different random seeds and reported all results with error bars. In \Cref{fig:thm1-delta} we can see that the radius of the cover (a smaller radius means the queries are more diverse) does have some variance due to the random initialization. However, the corresponding results in terms of detection accuracy in \Cref{fig:img_data} do have very low variance. Our interpretation is that for the CIFAR10 and F-MNIST experiments, the random initialization has little effect on detection performance.

\subsection{Results with Other Backbone Models}
\label{app:ablation_model}
\begin{table}[ht]
	\caption{$|\gQ|=20$. AUC ($\%$) with standard deviation for anomaly detection on six datasets (CIFAR-10, F-MNIST, Blood, OrganA, OrganC, OrganS). The backbone models are \gls{mhrot}~\citep{hendrycks2019using} and Deep SVDD~\citep{ruff2018deep}. For all experiments, we set the contamination ratio as $10\%$. \gls{aloe} consistently outperforms two best-performing baselines on all six datasets.}
	\label{tab:backbone-dsvdd-mhrot}
	\small
	\centering
	\vspace{1pt}
	
	\begin{tabular}{l@{\hskip 0.3in}cccc@{\hskip 0.3in}ccc}
        \toprule
        & \multicolumn{3}{c}{MHRot} & & 
        \multicolumn{3}{c}{Deep SVDD} \\
        \cmidrule(r){2-4} \cmidrule(r){6-8}
        & \gls{aloe} & \HybridA & \HybridB & & \gls{aloe} & \HybridA & \HybridB \\
		\cmidrule(r){2-4} \cmidrule(r){6-8}
		CIFAR-10 & \textbf{86.9$\pm$0.7} & 83.9$\pm$0.1 & 49.1$\pm$2.0 && \bf93.1$\pm$0.2 & 89.0$\pm$0.6 & 91.3$\pm$1.0 \\ 
		F-MNIST &\textbf{92.6$\pm$0.1} & 87.1$\pm$0.2 & 58.9$\pm$5.7 && \bf91.4$\pm$0.5 & 90.9$\pm$0.4 & 82.5$\pm$2.9\\
        Blood &\textbf{83.3$\pm$0.2} & 81.1$\pm$2.5 & 61.8$\pm$2.1 && \bf80.2$\pm$1.1 & 79.7$\pm$1.2 & 77.2$\pm$3.0\\
        OrganA & \textbf{96.5$\pm$0.3} & 94.1$\pm$0.3 & 61.1$\pm$4.8 && \bf89.5$\pm$0.3 & 87.1$\pm$0.7 & 71.3$\pm$3.8\\
        OrganC & \textbf{92.1$\pm$0.2} & 91.6$\pm$0.1 & 70.9$\pm$0.8  && \bf87.5$\pm$0.7 & 85.3$\pm$0.8 & 84.2$\pm$0.9\\
        OrganS & \textbf{89.3$\pm$0.2} & 88.3$\pm$0.3 & 68.2$\pm$0.1  && \bf85.5$\pm$0.7 & 83.4$\pm$0.3 & 81.2$\pm$1.3\\
        \bottomrule
	\end{tabular}
	
\end{table}
We are interested whether \gls{aloe} works for different backbone models. To that end, we repeat part of the experiments in \Cref{tab:img_results} but using an self-supervised learning model \gls{mhrot}~\citep{hendrycks2019using} and a one class classification model Deep SVDD~\citep{ruff2018deep} as the backbone model. We compare \gls{aloe} to two best performing baselines --- \HybridA~and \HybridB. In this experiment, \gls{mhrot} and Deep SVDD take different input types: while \gls{mhrot} takes raw images as input, Deep SVDD uses pre-trained image features. We also set the query budget to be $|\gQ|=20$.

We report the results in \Cref{tab:backbone-dsvdd-mhrot}. It showcases the superiority of \gls{aloe} compared to the baselines. On all datasets, \gls{aloe} significantly outperforms the two best performing baselines, \HybridA~and \HybridB, thus demonstrating the wide applicability of \gls{aloe} across anomaly detection model types.

\subsection{Robustness to Anomaly Ratios} 
\label{app:robust-anomaly-ratio}
\begin{figure*}[t!]
    \centering
    \includegraphics[width=0.5\linewidth]{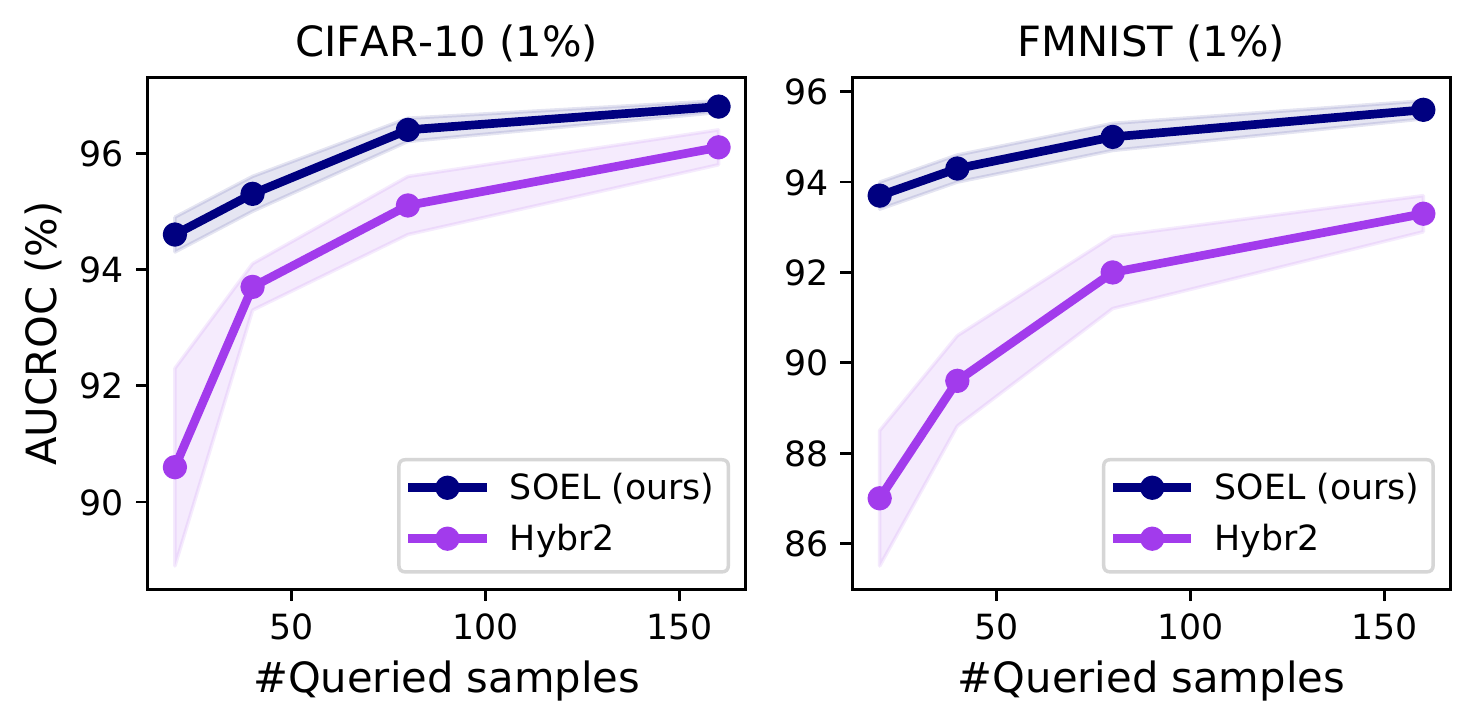}
    \includegraphics[width=0.5\linewidth]{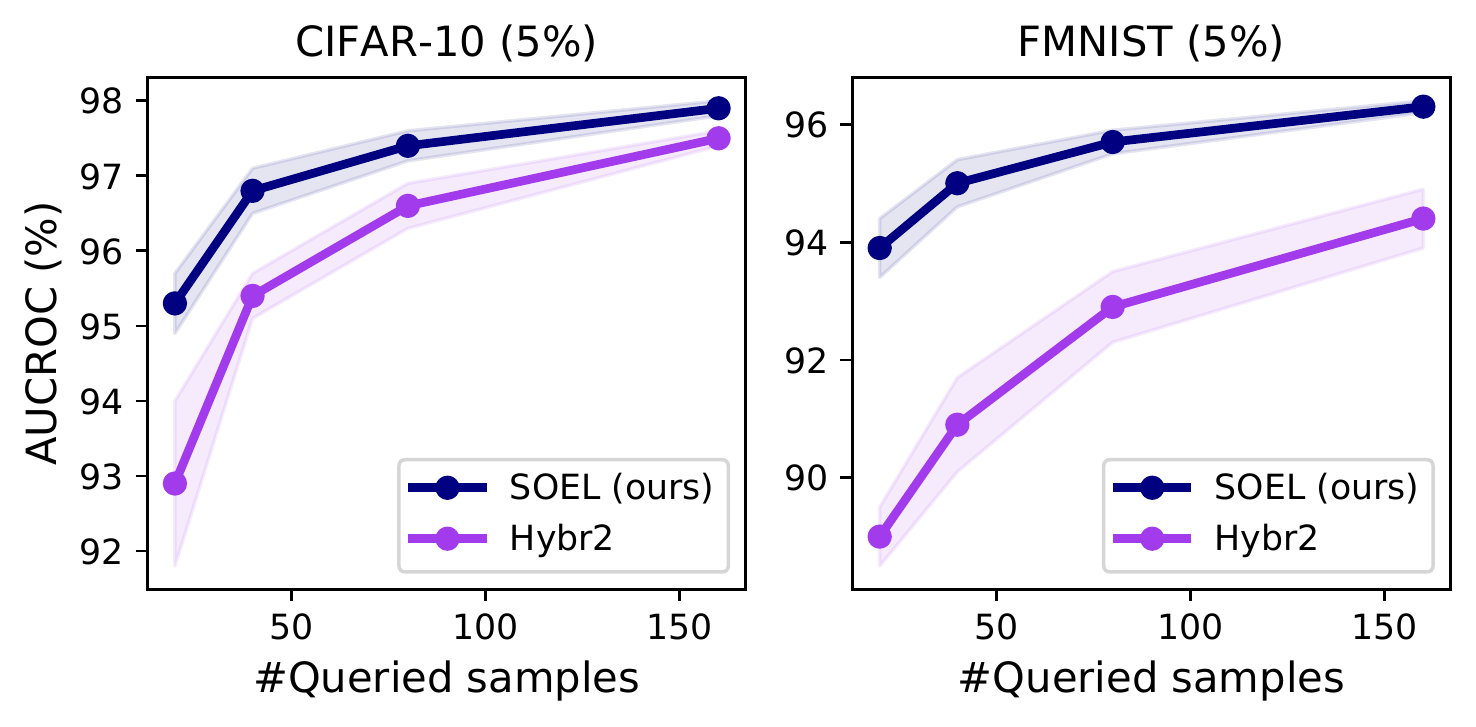}
    \includegraphics[width=0.5\linewidth]{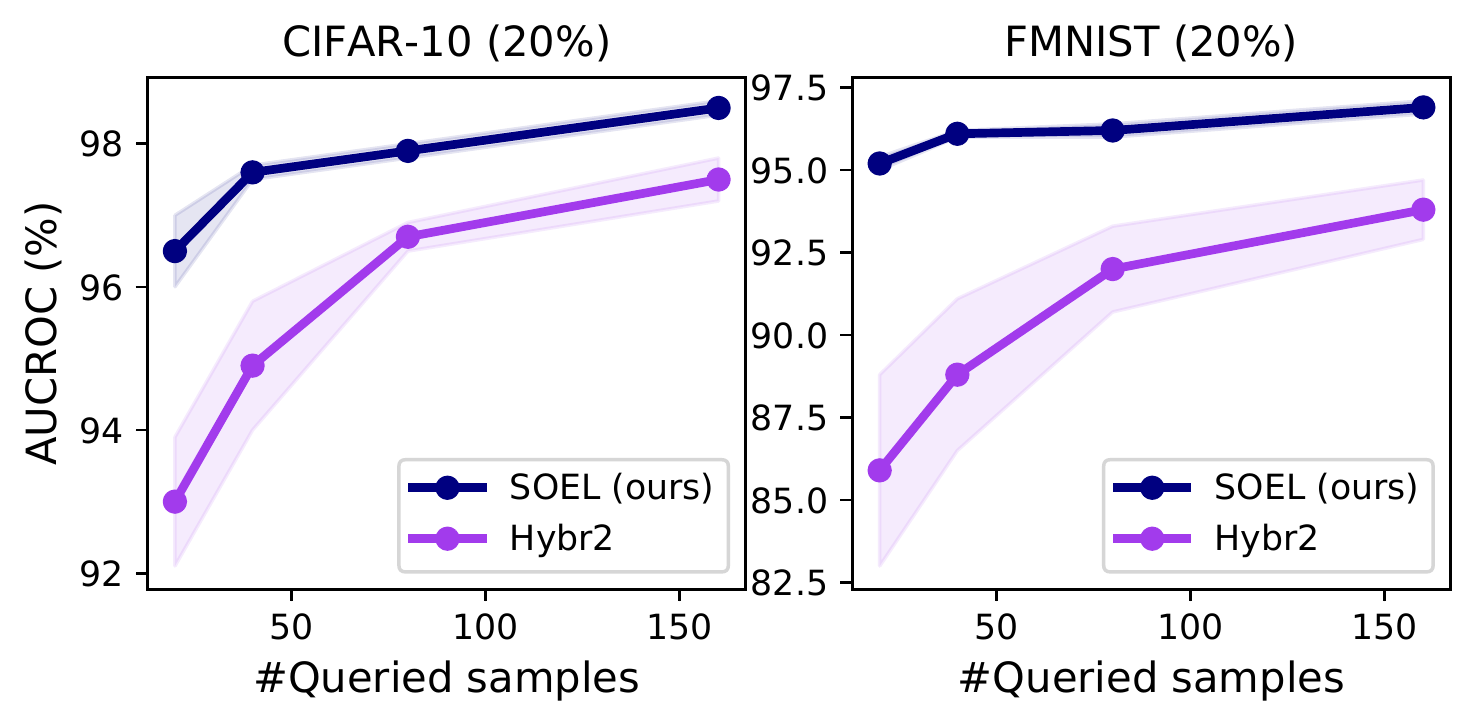}
    \vspace{-10pt}
    \caption{Running AUCs (\%) with different query budgets and data contamination ratios (1\%-top row, 5\%-middle row, 20\%-bottom row). Models are evaluated at $20, 40, 80, 160$ queries. \gls{aloe} performs the best on all three contamination ratio setups.}
    \label{fig:robust-ratio}
\vspace{-10pt}
\end{figure*}

Our method works for both low anomaly ratios and high anomaly ratios. In \Cref{fig:robust-ratio}, we compare \gls{aloe} against the best-performing baseline \HybridB on CIFAR-10 and FMNIST benchmarks. We vary the anomaly ratio among 1\%, 5\%, and 20\%. On all these three anomaly ratio settings, \gls{aloe} has significantly better performance than the baseline by over 2 percentage points on average.

\subsection{Disentanglement of \gls{aloe}} 
\label{app:ablation_component}
\begin{table}[ht]
\vspace{-5pt}
	\caption{$|\gQ|=20$. AUC ($\%$) with standard deviation for anomaly detection on CIFAR-10 and F-MNIST. For all experiments, we set the contamination ratio as $10\%$. \gls{aloe} mitigates the performance drop when \gls{ntl} and \gls{mhrot} trained on the contaminated datasets. Results of the unsupervised method LOE are borrowed from \citet{qiu2022latent}.}
	\label{tab:ablation-backbone}
	\small
	\centering
	\vspace{1pt}
	
	\begin{tabular}{l@{\hskip 0.3in}cccc@{\hskip 0.3in}ccc}
        \toprule
        & \multicolumn{3}{c}{NTL} & & 
        \multicolumn{3}{c}{MHRot} \\
        \cmidrule(r){2-4} \cmidrule(r){6-8}
        & LOE & \kmeans & \gls{aloe} & & LOE & \kmeans & \gls{aloe} \\
		\cmidrule(r){2-4} \cmidrule(r){6-8}
		CIFAR-10 & 94.9$\pm$0.1 &95.6$\pm$0.3 &\textbf{96.3$\pm$0.3}& &86.3$\pm$0.2 &64.0$\pm$0.2 &\textbf{86.9$\pm$0.7}  \\ 
		F-MNIST &92.5$\pm$0.1 &94.3$\pm$0.2 &\textbf{94.8$\pm$0.4}& &91.2$\pm$0.4 &91.5$\pm$0.1 &\textbf{92.6$\pm$0.1} \\
        \bottomrule
	\end{tabular}
	
\end{table}

We disentangle the benefits of each component of \gls{aloe} and compare it to unsupervised anomaly detection with latent outlier exposure (LOE) \citep{qiu2022latent}, and to supervised active anomaly detection with \kmeans querying strategy. Both active approaches (\kmeans and \gls{aloe}) are evaluated with $|\gQ|=20$ labeled samples. The unsupervised approach LOE requires an hyperparameter of the assumed data contamination ratio, which we set to the ground truth value $10\%$. Comparing \gls{aloe} to LOE reveals the benefits of the \kmeans active approach\footnote{
  Notice that while LOE uses the true contamination ratio (an oracle information), \gls{aloe} only uses the estimated contamination ratio by the 20 queries.
}; comparing \gls{aloe} to \kmeans reveals the benefits of the unsupervised loss function in \gls{aloe}. Results in \Cref{tab:ablation-backbone} show that \gls{aloe} leads to improvements for both ablation models.

\subsection{Comparison to Binary Classifier}
\label{app:bce}
\begin{figure*}[t!]
    \centering
    \includegraphics[width=1\linewidth]{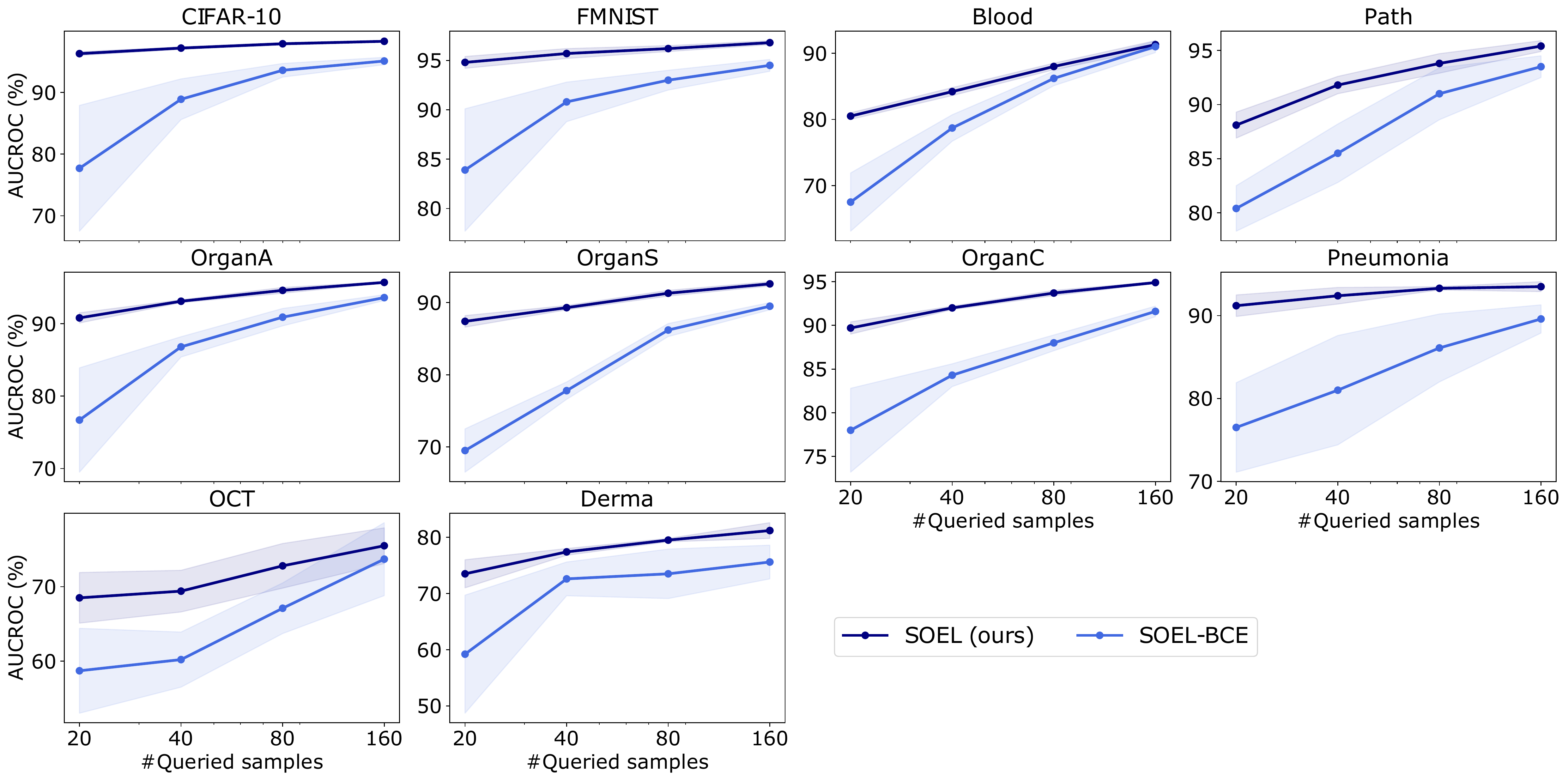}
    \vspace{-10pt}
    \caption{Running AUCs (\%) with different query budgets. Models are evaluated at $20, 40, 80, 160$ queries. Deep \gls{ad} model (\gls{ntl}) performs significantly better than a binary classifier.}
    \label{fig:img-data-bce}
\vspace{-10pt}
\end{figure*}

In the semi-supervised \gls{ad} setup, the labeled points can be seen as an imbalanced binary classification dataset. We, therefore, perform an ablation study where we only replace deep \gls{ad} backbone models with a binary classifier. All the other training and querying procedures are the same. We report the results on four different querying budget situations in \Cref{fig:img-data-bce}. The figure shows that a binary classifier on all 11 image datasets falls far short of the \gls{ntl}, a deep \gls{ad} model. The results prove that the inductive bias (learning compact representations for normal data) used by \gls{ad} models are useful for \gls{ad} tasks. However, such inductive bias is lacking for binary classifiers. Especially when only querying as few as 20 points, the model can't see all anomalies. The decision boundary learned by the classifier based on the queried anomalies possibly doesn't generalize to the unseen anomalies.

\subsection{Comparison to a Batch Sequential Setup}
\label{app:batch-seq}
\begin{figure*}[t!]
    \centering
    \includegraphics[width=0.7\linewidth]{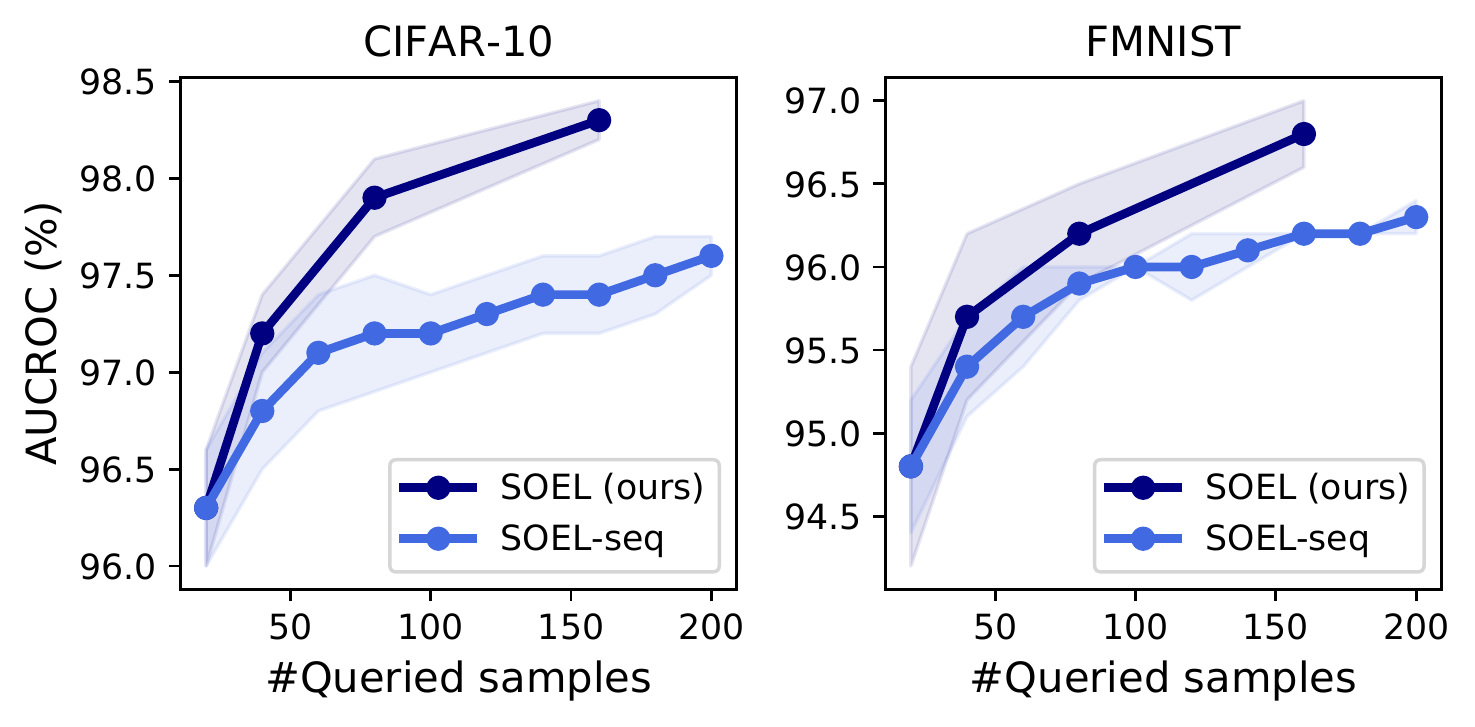}
    \vspace{-10pt}
    \caption{Running AUCs (\%) with different query budgets. Models are evaluated at $20, 40, 80, 160$ queries. \gls{aloe} performs better than a sequential version.}
    \label{fig:img-data-seq}
\vspace{-10pt}
\end{figure*}
In \Cref{fig:img-data-seq}, we extend our proposed method \gls{aloe} to a sequential batch active \gls{ad} setup. This sequential extension is possible because our querying strategy \kmeans is also a sequential one. At each round, we query 20 points and update the estimated contamination ratio. We plot this sequential version of \gls{aloe} and the original \gls{aloe} in \Cref{fig:img-data-seq} and make comparisons. The sequential version is not as effective as a single batch query of \gls{aloe}.

\subsection{Comparisons of Querying Strategies}
\label{app:ablation_query}
\begin{figure}[ht]
    \centering
    \includegraphics[width=1\linewidth]{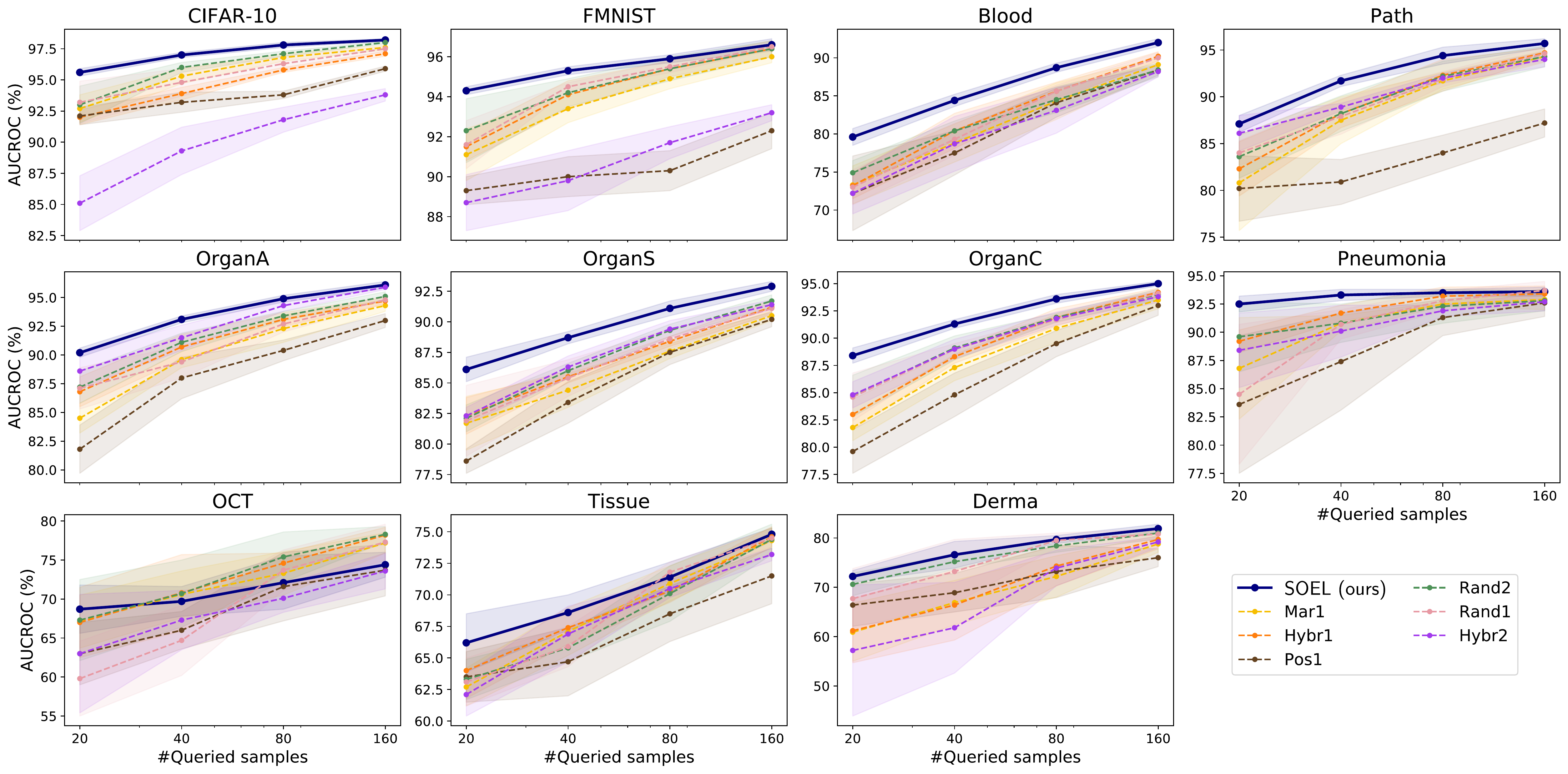}
    \caption{Ablation study on the query strategy. K-Means++ significantly outperforms other strategies for active anomaly detection on most of the datasets.} 
    \label{fig:ablation-query}
\end{figure}
To understand the benefit of sampling diverse queries with k-means++ and to examine the generalization ability (stated in \Cref{thm:rank}) of different querying strategies, we run the following experiment: We use a supervised loss on labeled samples to train various anomaly detectors. The only difference between them is the querying strategy used to select the samples. 
We evaluate them on all image data sets we study for varying number of queries $|\gQ|$ between $20$ and $160$. 

Results are in \Cref{fig:ablation-query}. 
On all datasets except OCT, k-means++ consistently outperforms all other querying strategies from previous work on active anomaly detection. The difference is particularly large when only few samples are queried. This also confirms that diverse querying generalizes better on the test data than other querying strategies (see additional results in \Cref{app:thm1}).

\subsection{Ablation on Estimated Contamination Ratio} 
\label{app:ablation_ratio}

\begin{figure}[ht]
    \centering
    \includegraphics[width=0.5\textwidth]{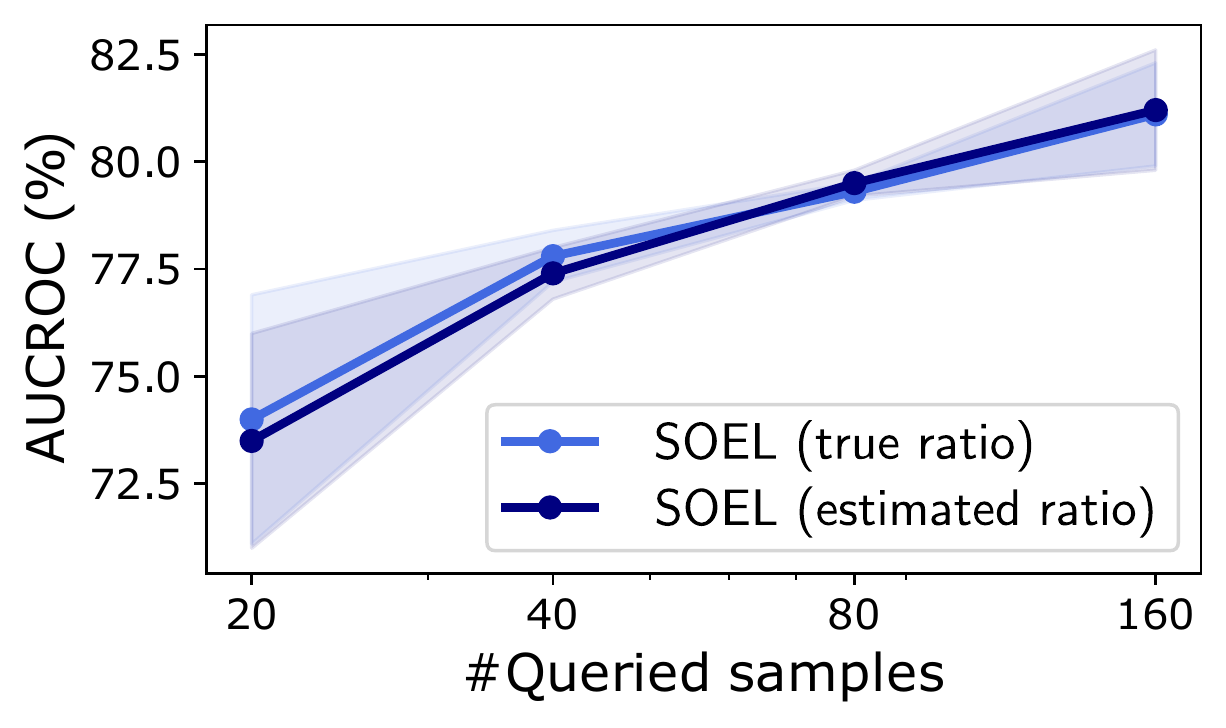}
    \vspace{-10pt}
    \caption{Model using the estimated ratio is indistinguishable from the one using the true ratio.} 
    \label{fig:ablation-true-est-ratio}
\end{figure}

To see how the estimated ratio affects the detection performance, we  compare \gls{aloe}  to the counterpart with the true anomaly ratio. We experiment on all 11 image datasets. In Fig.~\ref{fig:ablation-true-est-ratio}, we report the average results for all datasets when querying $|\gQ|=20, 40, 80, 160$ samples. It shows that \gls{aloe} with either true ratio or estimated ratio performs similar given all query budgets. Therefore, the estimated ratio can be applied safely. This is very important in practice, since in many applications the true anomaly ratio is not known.

\subsection{Ablations on Weighting Scheme} 
\label{app:ablation_weight}
\begin{figure}[ht]
    \centering
    \includegraphics[width=0.45\linewidth]{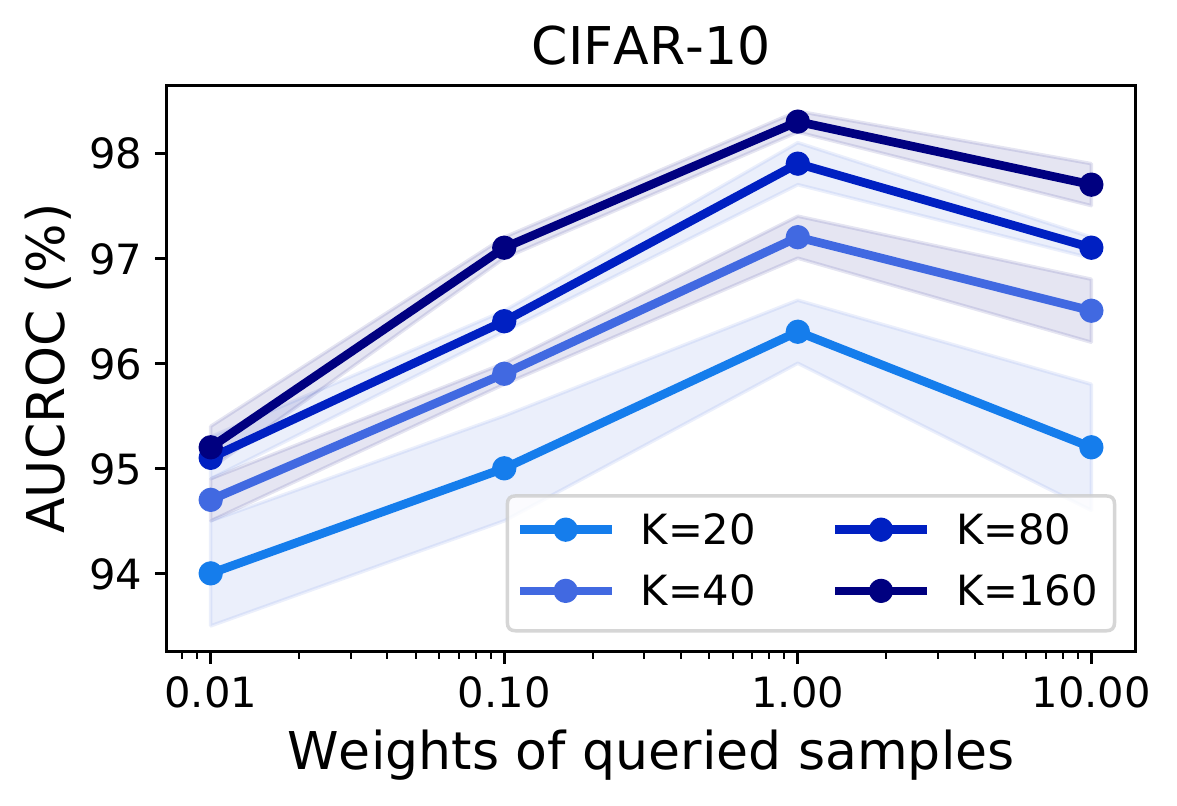}
    \includegraphics[width=0.45\linewidth]{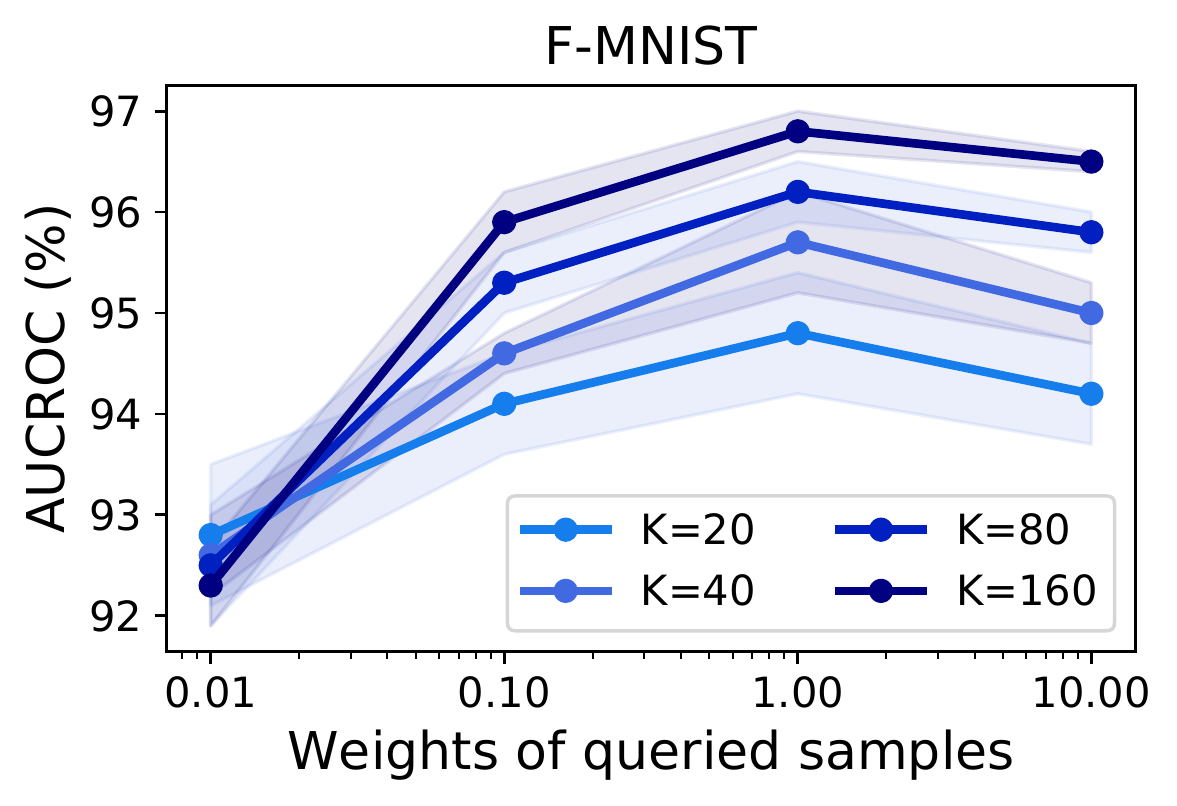}
    \caption{Ablation study on the weighting scheme in \Cref{eqn:loss-2}. With different query budgets $|\gQ|$, the performance on image datasets degrades both upon down-weighting (0.01, 0.1) or up-weighting (10.0) the queried samples. In contrast, equal weighting yields optimal results. } 
    \label{fig:ablation-weight}
\end{figure}
We make the implicit assumption that the \textit{averaged} losses over queried and unqueried data should be equally weighted (\Cref{eqn:loss-2}). That means, if a fraction $\epsilon$ of the data is queried, every queried data point weights $1/\epsilon$ as much as an unqueried datum. As a consequence, neither the queried nor the unqueried data points can dominate the result. 

To test whether this heuristic is indeed optimal, we added a scalar prefactor to the supervised loss in \Cref{eqn:loss-2} (the first term) and reported the results on the CIFAR-10 and F-MNIST datasets with different query budgets (\Cref{fig:ablation-weight}). A weight$<$1 corresponds to downweighting the queried term, while a weight$>$1 corresponds to upweighting it. We use the same experimental setup and backbone (NTL) as in the paper. The results are shown in \Cref{fig:ablation-weight}. We see that the performance degrades both upon down-weighting (0.01, 0.1) or up-weighting (10.0) the queried samples. In contrast, equal weighting yields optimal results. 

\subsection{Ablations on Temperature $\tau$}
\label{app:ablation_tau}
\begin{table}[ht]
\vspace{-5pt}
	\caption{Performance of ablation study on $\tau$. AUROCs (\%) on CIFAR-10 and F-MNIST when $|\gQ|=20$, the ground-truth contamination ratio is 0.1, and the backbone model is NTL. } %
	\label{tab:ablation-tau}
	\small
	\centering
	\vspace{1pt}
	\begin{tabular}{lcccc}
        \toprule
		   $\tau$  & 1 & 0.1 & 0.01 & 0.001  \\
		\midrule
		CIFAR-10 &$93.2 \pm 1.7$  & $94.5 \pm 0.8$  & $96.3 \pm 0.3$ & $95.9 \pm 0.4$  \\
		F-MNIST & $91.8 \pm 1.4$ & $92.7 \pm 1.1$ & $94.8 \pm 0.6$ & $94.9 \pm 0.2$ \\
        \bottomrule
	\end{tabular}
\end{table}
$\tau$ (in \Cref{eq:query-prob}) affects the querying procedure and smaller $\tau$ makes the querying procedure more deterministic and diverse because the softmax function (in \Cref{eq:query-prob}) can eventually become a maximum function. We add an ablation study on different values of $\tau$. We did experiments under the ground truth contamination ratio being 0.1 and $|Q|=20$. As \Cref{tab:ablation-tau} shows, the smaller $\tau$ results in better AUROC results (more diverse) and smaller errors (more deterministic).

\subsection{Ablations on Pseudo-label Values $\tilde y$}
\label{app:ablation-y}
\begin{table}[ht]
\vspace{-5pt}
	\caption{Performance of ablation study on $\tilde y$. AUROC (\%) on CIFAR-10 and F-MNIST when $|\gQ|=20$, the ground-truth contamination ratio is 0.1, and the backbone model is NTL. } %
	\label{tab:ablation-pseudo-y}
	\small
	\centering
	\vspace{1pt}
	\begin{tabular}{lcccccc}
        \toprule
		$\tilde y$ & 1.0 &0.875  &0.75 &0.625 &0.5 &0.25   \\
		\midrule
		CIFAR-10 &95.3 $\pm$ 0.6  &95.7 $\pm$ 0.4  &95.8 $\pm$ 0.4 &96.0 $\pm$ 0.5 &96.3 $\pm$ 0.3 &94.5 $\pm$ 0.3  \\
		F-MNIST & 94.5  $\pm$ 0.5  &94.5 $\pm$ 0.4  &94.6 $\pm$ 0.4 &94.6 $\pm$ 0.3 &94.8 $\pm$ 0.6 &94.0 $\pm$ 0.4\\
        \bottomrule
	\end{tabular}
\end{table}
Analyzing the effects of the pseudo-label values $\tilde y$ is an interesting ablation study. Therefore, we perform the following experiments to illustrate the influence of different $\tilde y$ values.  We set the ground truth contamination ratio being 0.1 and $|Q|=20$. We vary the $\tilde y$ from 0.25 to 1.0 and conduct experiments. For each $\tilde y$ value, we run 5 experiments with different random seeds and report the AUROC results with standard deviation. It shows that $\tilde y=0.5$ performs the best. While the performance of CIFAR-10 degrades slightly as $\tilde y$ deviates from 0.5, F-MNIST is pretty robust to $\tilde y$. All tested $\tilde y$ outperform the best baseline reported in \Cref{tab:img_results}.

\subsection{Comparisons with Semi-supervised Learning Frameworks} 
\label{app:ablation_labeling}
\begin{figure}[ht]
    \centering
    \includegraphics[width=0.45\linewidth]{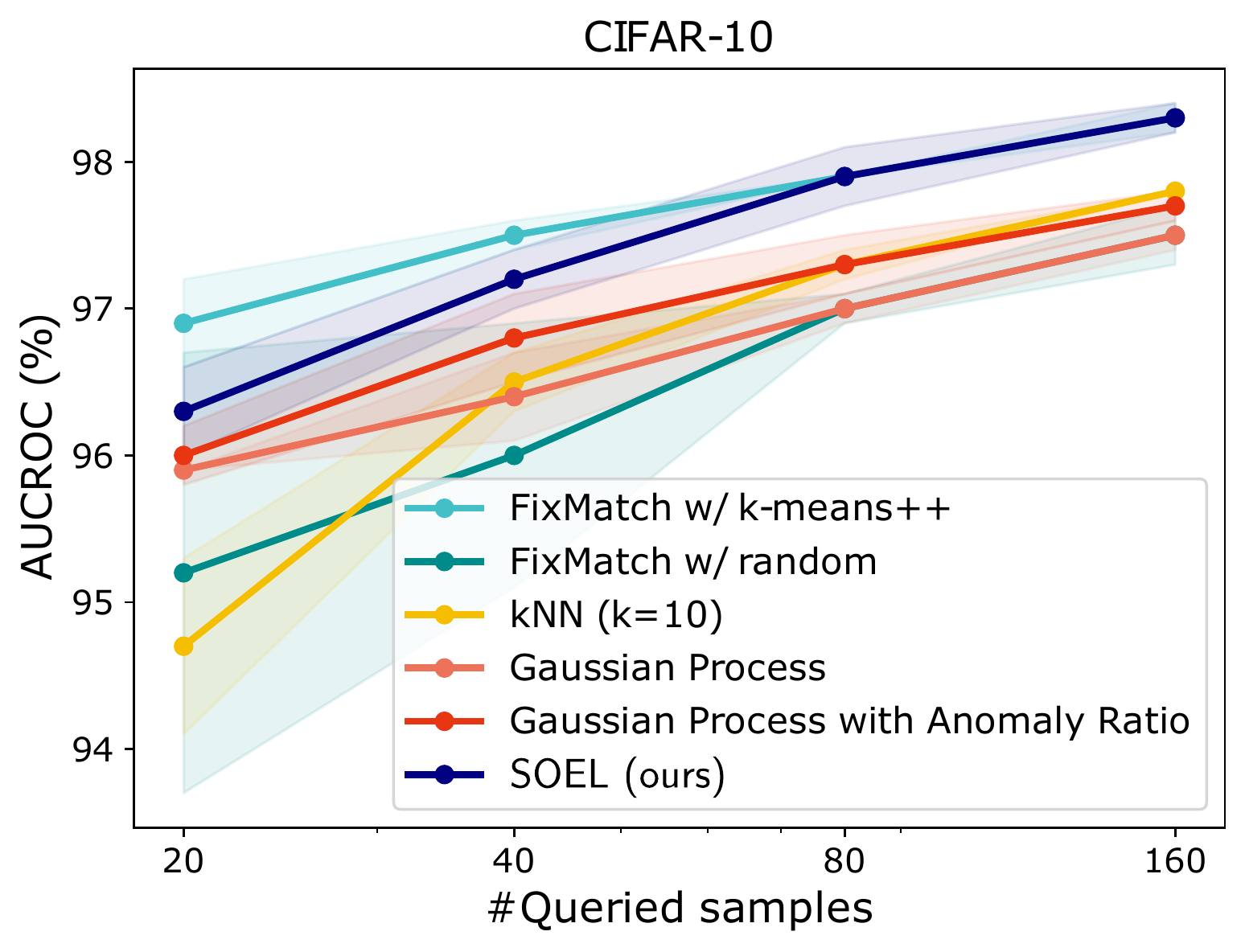}
    \includegraphics[width=0.45\linewidth]{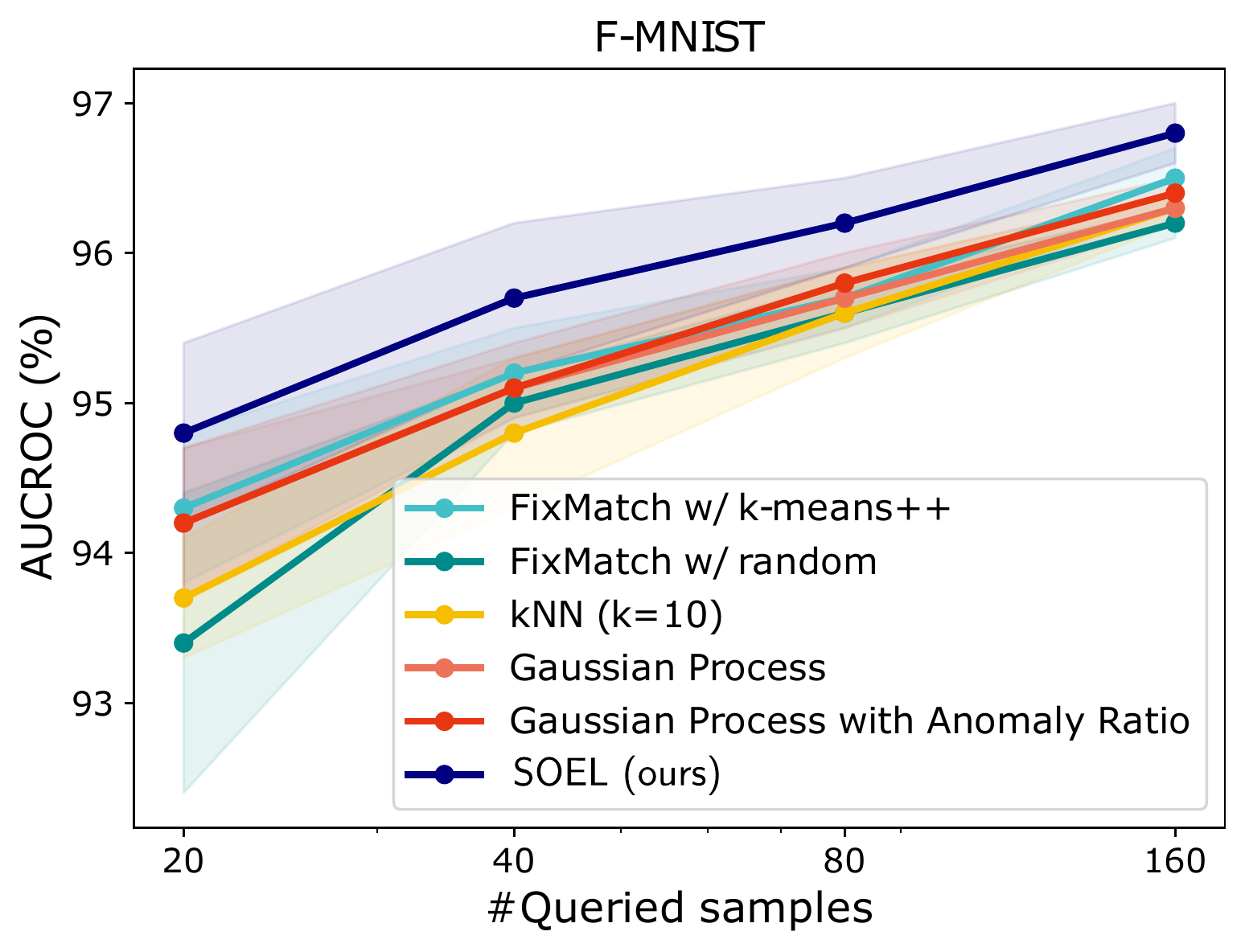}
    \caption{Comparison with semi-supervised learning fraemworks, FixMatch \citep{sohn2020fixmatch}, $k$-nearest neighbors \citep{iscen2019label}, and Gaussian process \citep{li2018pseudo}. On F-MNIST, \gls{aloe} outperforms all baselines, while on CIFAR-10, \gls{aloe} has a comparable performance with FixMatch with \kmeans querying.} 
    \label{fig:semisuper-comparison}
\end{figure}
\gls{aloe} exploits the unlabeled data to improve the model performance. This shares the same spirit of semi-supervised learning. We are curious about how a semi-supervised learning method performs in our active anomaly detection setup. To this end, we adapted an existing semi-supervised learning framework FixMatch \citep{sohn2020fixmatch} to our setup and compared with our method in \Cref{fig:semisuper-comparison}. As follows, we will first describe the experiment results and then state the adaptation of FixMatch to anomaly detection we made. 

FixMatch, as a semi-supervised learning algorithm, regularizes the image classifier on a large amount of unlabeled data. The regularization, usually referred to consistency regularization, requires the classifier to have consistent predictions on different views of unlabeled data, thus improves the classifier's performance. FixMatch generates various data views through image augmentations followed by Cutout \citep{devries2017improved}. We noticed that, although FixMatch focuses on making use of the unlabeled data, its performance is highly affected by the quality of the labeled data subset. We investigated two variants depending on how we acquire the labeled data. One is the original semi-supervised learning setting, i.e., assuming the labeled data is a random subset of the whole dataset. The other one utilizes the same diversified data querying strategy \kmeans as \gls{aloe} to acquire the labeled part. In \Cref{fig:semisuper-comparison}, we compared the performance of the two variants with \gls{aloe}. It shows that, on natural images CIFAR10 for which FixMatch is developped, while the original FixMatch with random labeled data is still outperformed by \gls{aloe}, FixMatch with our proposed querying strategy \kmeans has a comparable performance with \gls{aloe}. However, such advantage of FixMatch diminishes for the gray image dataset F-MNIST, where both variants are beat by \gls{aloe} on all querying budgets.  
In addition, the FixMatch framework is restrictive and may not be applicable for tabular data and medical data, as the augmentations are specially designed for natural images. 

FixMatch is designed for classification. To make it suit for anomaly detection, we adapted the original algorithm\footnote{
  We adapted the FixMatch implementation \url{https://github.com/kekmodel/FixMatch-pytorch}
} and adopted the following procedure and loss function.
\begin{enumerate}[leftmargin=*,itemsep=0pt]
    \item Label all training data as normal and train the anomaly detector for one epoch;
    \item Actively query a subset of data with size $|\gQ|$, resulting in $\gQ$ and the remaining data $\gU$;
    \item Finetune the detector in a supervised manner on non-augmented $\gQ$ for 5 epochs;
    \item Train the detector with the FixMatch loss \Cref{eq:fixmatch-loss} on augmented $\{\gU, \gQ\}$ until convergence.
\end{enumerate}

We denote weak augmentation of input $\vx$ by $\alpha(\vx)$ and the strong augmentation by $\gA(\vx)$. The training objective function we used is
\begin{align}
\label{eq:fixmatch-loss}
    &~{\cal L}_{\text{FixMatch}}(\theta) = \frac{1}{|\gQ|}\sum_{j\in \gQ} \big(y_j \La(\alpha(\vx_j)) + (1-y_j) \Ln(\alpha(\vx_j)) \big) \nonumber\\ &~+ \frac{1}{|\gU|}\sum_{i\in \gU} \I(S(\alpha(\vx_i))<q_{0.7}\text{ or }S(\alpha(\vx_i))>q_{0.05}) \big(\tilde{y}_i \La(\gA(\vx_i)) + (1-\tilde{y}_i) \Ln(\gA(\vx_i))\big)
\end{align}
where pseudo labels $\tilde{y}_i=\I(S(\alpha(\vx_i))>q_{0.05}))$ and $q_{n}$ is the $n$-quantile of the anomaly scores $\{S(\alpha(\vx_i))\}_{i\in\gU}$. In the loss function, we only use the unlabeled samples with confidently predicted pseudo labels. This is controlled by the indicator function $\I(S(\alpha(\vx_i))<q_{0.7}\text{ or }S(\alpha(\vx_i))>q_{0.05})$. We apply this loss function for mini-batches on a stochastic optimization basis.

We also extend the semi-supervised learning methods using non-parametric algorithms to our active anomaly detection framework. We applied $k$-nearest neighbors and Gaussian process for inferring the latent anomaly labels \citep{iscen2019label,li2018pseudo} because these algorithms are unbiased in the sense that if the queried sample size is large enough, the inferred latent anomaly labels approach to the true anomaly labels.  For these baselines, we also queried a few labeled data with \kmeans-based diverse querying strategy and then annotate the unqueried samples with k-nearest neighbor classifier or Gaussian process classifer trained on the queried data.

Both methods become ablations of \gls{aloe}. We compare \gls{aloe} with them on CIFAR-10 and F-MNIST under various query budgets and report their results in \Cref{fig:semisuper-comparison}. 
On both datasets, \gls{aloe} improves over the variant of using only queried samples for training. 
On F-MNIST, \gls{aloe} outperforms all ablations clearly under all query budgets, while on CIFAR-10, \gls{aloe} outperforms all ablations except for FixMatch when query budget is low. 
In conclusion, \gls{aloe} boosts the performance by utilizing the unlabeled samples properly, while other labeling strategies are less effective.

\subsection{More Comparisons}
\label{app:more-comp}
\begin{table}[ht]
\vspace{-5pt}
    \caption{Comparisons with kNN method. We reported the F1-score ($\%$) with standard error for anomaly detection on tabular datasets when the query budget $K=10$. \gls{aloe} outperforms the kNN baseline. }
\label{tab:tab_knn}
    \centering
    \small
  \begin{tabular}{l|cc}
  \toprule
     & \bfseries $k^{\text{th}}$NN & \bfseries ALOE \\ 
  \midrule
\bfseries  BreastW  & 92.5$\pm$2.1	& \textbf{93.9$\pm$0.5}	\\ 
\bfseries Ionosphere & 88.1$\pm$1.3	& \textbf{91.8$\pm$1.1}	\\ 
 \bfseries Pima & 40.5$\pm$4.7	& \textbf{55.5$\pm$1.2}	\\ 
 \bfseries Satellite & 61.1$\pm$2.2	& \textbf{71.1$\pm$1.7}	\\ 
\midrule
\bfseries Average & 70.6 & \textbf{78.1} \\
  \bottomrule 
  \end{tabular}
 \vspace{-5pt}
\end{table}

\begin{figure}[ht]
    \centering
    \includegraphics[width=0.45\linewidth]{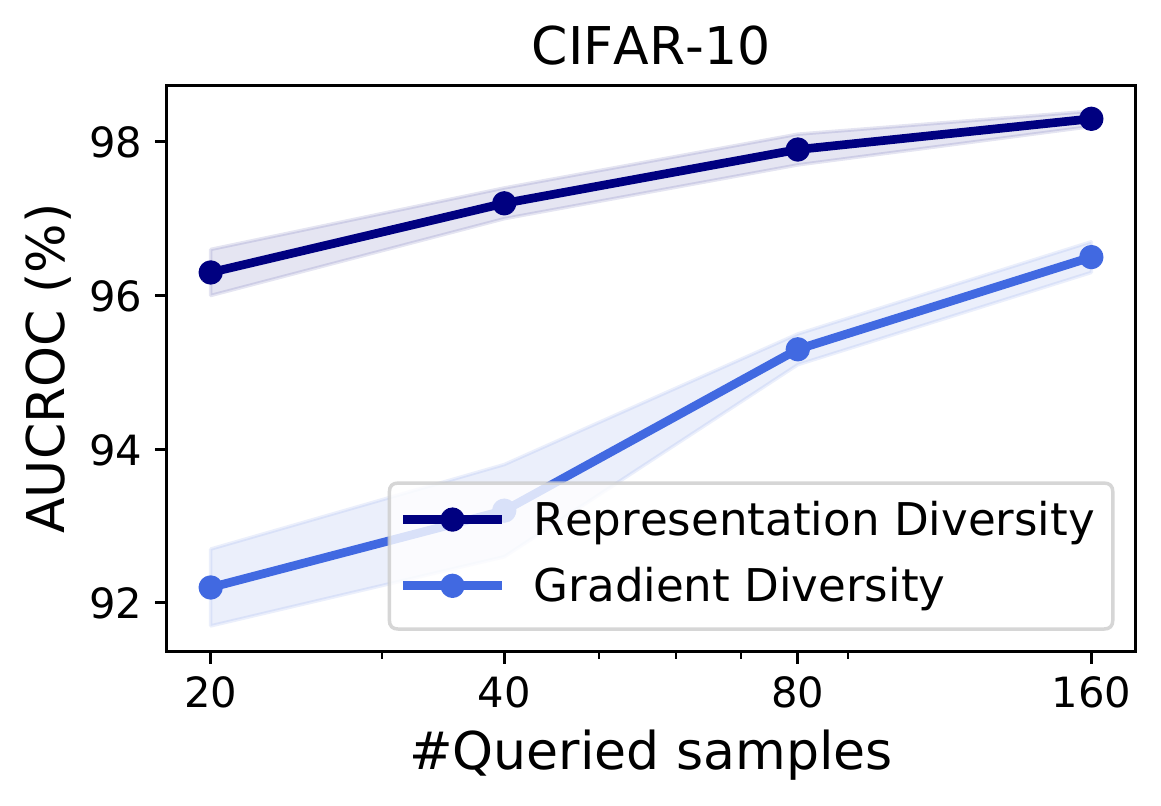}
    \includegraphics[width=0.45\linewidth]{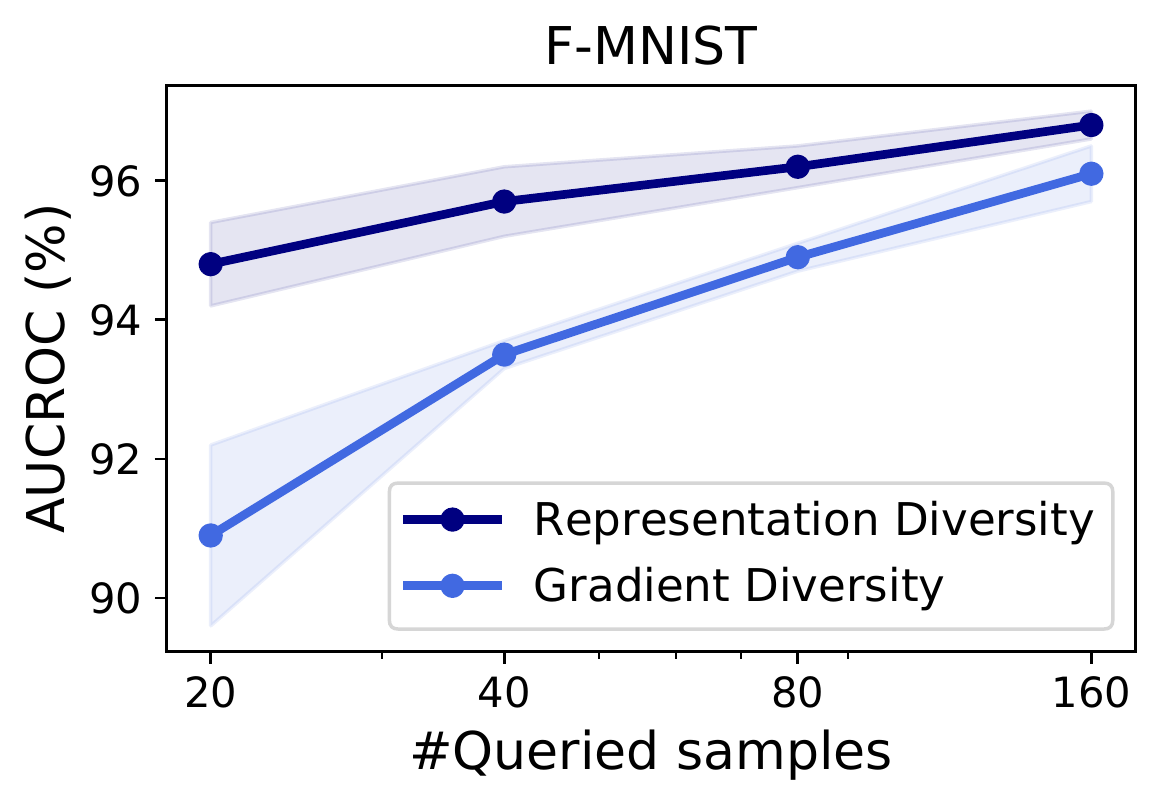}
    \caption{Comparison with gradient diversity querying strategy (BADGE)~\citep{ash2020deep}. The gradients wrt. the penultimate layer representation don't provide as informative queries as the representation itself, thus outperformed by our querying strategy \gls{aloe}. The true contamination ratio is 10\%.} 
    \label{fig:grad-div-comparison}
\end{figure}
\paragraph{Comparisons to kNN~\citep{ramaswamy2000efficient}}
We compared against kNN in two ways. First we confirmed that our baseline backbone model NTL is competitive with kNN, which is shown to have a strong performance on tabular data~\citep{shenkar2022anomaly}. To this end, NTL has been shown to yield 95.7\% AUC on clean CIFAR-10 data, see \citealp[Table 1 column 1]{shenkar2022anomaly}. In contrast, \citet{qiu2022latent} reported 96.2\% AUC in Table 2, which is very close.  

Second, we tested the performance of the kNN method on our corrupted training data set. We gave kNN the advantage of using the ground truth contamination ratio (otherwise when under-estimating this value, we saw the method degrade severely in performance). 

KNN has two key hyperparameters: the number of nearest neighbors $k$ and the assumed contamination ratio of the training set. The method uses this assumed contamination ratio when fitting to define the threshold on the decision function. In our experiments, we tried multiple values of $k$ and reported the best testing results. Although the ground truth anomaly rate is unknown and our proposed methods don’t have access to it, we gave kNN the competitive advantage of knowing the ground truth contamination ratio. 

We studied the same tabular data sets as in our paper: BreastW, Ionosphere, Pima, and Satellite. We used the same procedure for constructing contaminated data outlined in our paper, where the contamination ratio was set to 10\%. The results are summarized in \Cref{tab:tab_knn}. 

We adopted PyOD’s implementation of kNN\footnote{
  \url{https://github.com/yzhao062/pyod}
} and set all the other hyperparameters to their default values (“method, radius, algorithm, leaf\_size, metric, p, and metric\_params”). We repeated the experiments 10 times and reported the mean and standard deviation of the F1 scores in \Cref{tab:tab_knn}. We find that our active learning framework outperforms the kNN baseline. 

In more detail, the F1 scores for different values of $k$ are listed below, where $k = 1, 2, 5, 10, 15, 20$, respectively:
\begin{itemize}
    \item BreastW: 84.3$\pm$7.6, 86.5$\pm$3.1, 89.9$\pm$3.9, 90.7$\pm$3.4, 92.5$\pm$2.1, 91.9$\pm$1.5
    \item Ionosphere: 88.1$\pm$1.3, 87.6$\pm$2.6, 84.5$\pm$3.9, 75.2$\pm$2.5, 70.4$\pm$3.6, 67.4$\pm$3.4
    \item Pima: 34.4$\pm$3.6, 32.3$\pm$3.4, 36.9$\pm$6.4, 40.5$\pm$4.7, 35.3$\pm$3.6, 35.5$\pm$4.5
    \item Satellite: 51.0$\pm$1.1, 53.5$\pm$0.7, 54.7$\pm$1.3, 57.4$\pm$1.8, 59.3$\pm$1.3, 61.1$\pm$2.2
\end{itemize}

\paragraph{Comparisons to Gradient Diversity Querying Strategy (BADGE)~\citep{ash2020deep}} We compared against a popular active learning method, BADGE~\citep{ash2020deep}, which is a diversity-driven active learning method that exploits sample-wise gradient diversity. We start with observing that BADGE doesn't work well for anomaly detection in \Cref{fig:grad-div-comparison}, where we only replaced the objects that \kmeans works on in \gls{aloe} with gradients demanded in BADGE~\citep{ash2020deep} while keeping all other settings fixed. This variant is referred to as "Gradient Diversity" while ours is denoted by "Representation Diversity". \Cref{fig:grad-div-comparison} shows the performance of Gradient Diversity is outperformed by a large margin, failing in querying informative samples as our Representation Diversity. 

To understand which part of BADGE breaks for anomaly detection tasks, we check the gradients used by BADGE in an anomaly detection model. Before that, we start with describing how BADGE works. BADGE is developed for active learning in classification tasks. Given a pre-trained classifier, it first predicts the most likely label $\hat y$ (pseudo labels) for the unlabeled training data $\vx$. These pseudo labels are then used to formulate a cross entropy loss $l_{CE}(\vx, \hat y)$. BADGE computes every data point's loss function's gradient to the final layer's weights as the data's representation. Upon active querying, a subset of data are selected such that their representations are diverse. In particular, the gradient to each class-specific weight $W_k$ is $\nabla_{W_k} l_{CE}(\vx, \hat y) = (p_k - \I(\hat y=k))\phi(\vx)$ where $p_k$ is the predicted probability of being class $k$ and $\phi(\vx)$ is the output of the penultimate layer. Proposition 1 of \citet{ash2020deep} shows the norm of the gradient with pseudo labels is a lower bound of the one with true labels.
In addition, note that the gradient is a scaling of the penultimate layer output. The scaling factor describes the predictive uncertainty and is upper bounded by 1. Therefore, the gradients are informative surrogates of the penultimate layer output of the network, as shown by the inequality
\begin{align}
    ||\nabla_{W_k} l_{CE}(\vx, \hat y)||^2 \leq ||\nabla_{W_k} l_{CE}(\vx, y)||^2 \leq ||\phi(\vx)||^2.
\end{align}
However, these properties are associated with the softmax activation function usage. In anomaly detection, models and losses are diverse and are beyond the usage of softmax activation outputs. Hence the gradients are no longer good ways to construct active queries. For example, the supervised deep SVDD~\citep{ruff2019deep} uses the contrasting loss $l(\vx, y)=y/(W\phi(\vx) - \vc)^2 + (1-y)(W\phi(\vx) - \vc)^2$ to compact the normal sample representations around center $\vc$. However, the gradient $\nabla_W l(\vx, y)=\big(2(1-y)(W\phi(\vx)-c) - 2y(W\phi(\vx)-c)^{-3}\big)\phi(\vx)$ is not a bounded scaling of $\phi(\vx)$ any more, thus not an informative surrogate of point $\vx$.

\subsection{NTL as a Unified Backbone Model}
\label{sec:original_baseline}

In Section 4 of the main paper, we have empirically compared \gls{aloe} to active-learning strategies known from various existing papers, where these strategies originally were proposed using different backbone architectures (either shallow methods or simple neural architectures, such as autoencoders). However, several recent benchmarks have revealed that these backbones are no longer competitive with modern self-supervised ones \citep{alvarez2022revealing}. For a fair empirical comparison of \gls{aloe} to modern baselines, we upgraded the previously proposed active-learning methods by replacing their simple respective backbones with a modern self-supervised backbone: \gls{ntl} \citep{qiu2021neural}---the same backbone that is also used in \gls{aloe}.

\begin{figure}[ht]
    \centering
    \includegraphics[width=0.85\linewidth]{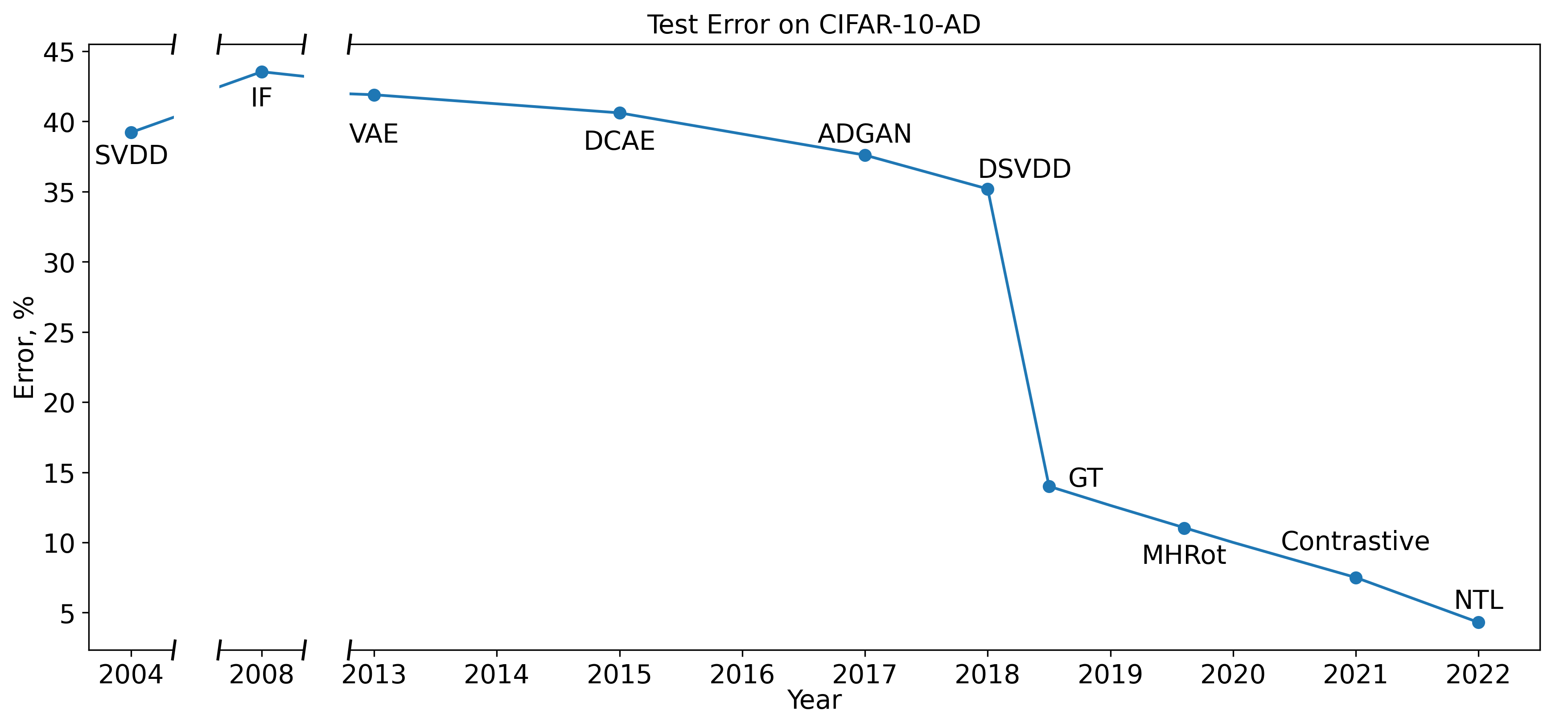}
  \caption{Error (in $\%$ of 1-AUCROC) of ten methods on CIFAR10: two shallow methods (SVDD \citep{TaxD04} and IF \citep{liu2008isolation}) and eight deep methods (VAE \citep{kingma2014vae}, DCAE \citep{makhzani2015winner}, ADGAN \citep{deecke2018image}, DSVDD \citep{ruff2018deep}, GT \citep{golan2018deep}, MHRot \citep{hendrycks2019using}, Contrastive \citep{sohn2020learning}, and NTL \citep{qiu2022latent}). \gls{ntl} achieves the best anomaly detection performance on CIFAR10.
    \label{fig:image_benchmark}}
\end{figure}

We motivate our choice of \gls{ntl} as unified backbone in our experiments as follows.  \Cref{fig:image_benchmark} shows the results of ten shallow and deep anomaly detection methods  \citep{TaxD04,liu2008isolation,kingma2014vae,makhzani2015winner,deecke2018image,ruff2018deep,golan2018deep, hendrycks2019using,sohn2020learning,qiu2022latent} on the CIFAR10 one-vs.-rest anomaly detection task. NTL performs best (by a large margin) among the compared methods, including many classic backbone models known from the active anomaly detection literature \citep{gornitz2013toward,barnabe2015active,das2019active,ruff2019deep,pimentel2020deep,trittenbach2021overview,ning2022deep}. 

\begin{table}[ht]
    \caption{F1-scores (in $\%$) and their standard deviations of 13 anomaly detection methods on tabular data. Results are taken from \citet{alvarez2022revealing}. The results indicate that NTL is the state-of-the-art for tabular anomaly detection.}
\label{tab:tab_benchmark}
 	\small
    \centering
  \begin{tabular}{l|ccccc|c}
  \toprule
     & KDDCUP10 & NSL-KDD & IDS2018& Arrhythmia & Thyroid & Avg.\\ 
  \midrule
ALAD & 95.9$\pm$0.7 & 92.1$\pm$1.5 & 59.0$\pm$0.0 & 57.4$\pm$0.4& 68.6$\pm$0.5 & 74.6  \\         
DAE & 93.2$\pm$2.0 &\bf96.1$\pm$0.1 &\bf71.5$\pm$0.5 & 61.5$\pm$2.5 & 59.0$\pm$1.5 & 76.3 \\
DAGMM & 95.9$\pm$1.4 &85.3$\pm$7.4 &55.8$\pm$5.3 & 50.6$\pm$4.7 & 48.6$\pm$8.0 & 67.2 \\
DeepSVDD &89.1$\pm$2.0 &89.3$\pm$2.0 &20.8$\pm$11 & 55.5$\pm$3.0&  13.1$\pm$13 & 53.6 \\
DROCC & 91.1$\pm$0.0 &90.4$\pm$0.0 &45.6$\pm$0.0 & 35.8$\pm$2.6 & 62.1$\pm$10 & 65.0 \\
DSEBM-e & 96.6$\pm$0.1 &94.6$\pm$0.1 &43.9$\pm$0.8 & 59.9$\pm$1.0 & 23.8$\pm$0.7 & 63.8 \\
DSEBM-r & \bf98.0$\pm$0.1 &95.5$\pm$0.1 &40.7$\pm$0.1 & 60.1$\pm$1.0 & 23.6$\pm$0.4 & 63.6 \\
DUAD & 96.5$\pm$1.0 &94.5$\pm$0.2 &\bf71.8$\pm$2.7 & 60.8$\pm$0.4 & 14.9$\pm$5.5 & 67.7 \\
MemAE &95.0$\pm$1.7 &95.6$\pm$0.0 &59.9$\pm$0.1 & 62.6$\pm$1.6& 56.1$\pm$0.9 & 73.8 \\
SOM-DAGMM &97.7$\pm$0.3 &95.6$\pm$0.3 &44.1$\pm$1.1 & 51.9$\pm$5.9 & 52.7$\pm$12 & 68.4 \\
LOF &95.1$\pm$0.0 &91.1$\pm$0.0 &63.8$\pm$0.0 & 61.5$\pm$0.0 & 68.6$\pm$0.0 & 76.0 \\
OC-SVM &96.7$\pm$0.0 &93.0$\pm$0.0 &45.4$\pm$0.0 & \bf63.5$\pm$0.0 & 68.1$\pm$0.0 &  73.3 \\
NTL &96.4$\pm$0.2 &\bf96.0$\pm$0.1 &59.5$\pm$8.9 & 60.7$\pm$3.7 & \bf73.4$\pm$0.6 & \bf77.2 \\
  \bottomrule 
  \end{tabular}
\end{table}

An independent benchmark comparison of 13 methods (including nine deep methods proposed in 2018--2022) \citep{alvarez2022revealing}  recently identified NTL as the leading anomaly-detection method on tabular data. In their summary, the authors write: 'NeuTraLAD, the transformation-based approach, offers consistently above-average performance across all datasets. The data-augmentation strategy is particularly efficient on small-scale datasets where samples are scarce.'. Note that the latter is also the scenario where active learning is thought to be the most promising. We show the results from \citet{alvarez2022revealing} in \Cref{tab:tab_benchmark}.

\end{document}